\documentclass[letterpaper, 10 pt, conference]{ieeeconf}  

\IEEEoverridecommandlockouts                              

\usepackage{graphicx}
\usepackage{comment}
\usepackage{bm, amsmath,amssymb} 
\usepackage{breqn}
\usepackage{color}
\usepackage{algorithm}	
\usepackage{algorithmic}

\usepackage{psfrag}

\title{\LARGE \bf
Revisiting visual-inertial structure from motion \\for odometry and SLAM initialization
}

\author{Georgios Evangelidis ~~~~~ Branislav Micusik\\
Snap Inc.\\
Vienna, Austria\\
{\tt\small georgios@snap.com~~~~~brano@snap.com}
}

\usepackage{fancyhdr}
\fancypagestyle{mystyle}{%
    \fancyhead[L]{A compressed version will appear in IEEE Robotics \& Automation Letters, 2021}\fancyfoot{}
}%
\begin{document}

\newcommand{\Figs}{Figs/}

\newcommand{\etal}{\mbox{\emph{et al.\ }}}
\newcommand{\ie}{\mbox{\emph{i.e.\ }}}
\newcommand{\wrt}{\mbox{\emph{w.r.t.\ }}}
\newcommand{\eg}{\mbox{\emph{e.g.\ }}}
\newcommand{\etc}{\mbox{\emph{etc}}}

\newcommand{\Eq}[1]{Eq.\,(\ref{#1})}  
\newcommand{\Fig}[1]{Fig.\,\ref{#1}}  
\newcommand{\Sec}[1]{Sec.\,\ref{#1}}  
\newcommand{\Tab}[1]{Tab.\,\ref{#1}}  

\newcommand{\M}[1]{\mathtt{#1}}   
\newcommand{\V}[1]{\mathbf{#1}}   
\newcommand{\N}[1]{\mathbb{#1}}   
\newcommand{\C}[1]{\mathcal{#1}}  

\newcommand{\MM}[1]{\textsc{#1}}  
\newcommand{\dist}{\mathrm{dist}}
\newcommand{\imu}{\textsc{imu}}

\maketitle
\thispagestyle{mystyle}
\pagestyle{empty}

\begin{abstract}
In this paper, an efficient closed-form solution for the state initialization in visual-inertial odometry (VIO) and simultaneous localization and mapping (SLAM) is presented. Unlike the state-of-the-art, we do not derive linear equations  from triangulating pairs of point observations. Instead, we build on a direct triangulation of the unknown $3D$ point paired with each of its observations. We show and validate the high impact of such a simple difference. The resulting linear system has a simpler structure and the solution through analytic elimination only requires solving a $6\times 6$ linear system (or $9 \times 9$ when accelerometer bias is included). In addition, all the observations of every scene point are jointly related, thereby leading to a less biased and more robust solution. The proposed formulation attains up to $50$ percent decreased velocity and point reconstruction error compared to the standard closed-form solver, while it is $4\times$ faster for a $7$-frame set. Apart from the inherent efficiency, fewer iterations are needed by any further non-linear refinement thanks to better parameter initialization. In this context, we provide the analytic Jacobians for a non-linear optimizer that optionally refines the initial parameters. The superior performance of the proposed solver is established by quantitative comparisons with the state-of-the-art solver. 
\end{abstract}

\section{Introduction}\label{sec:Introduction}
Visual odometry~\cite{Nister-CVPR2005} or SLAM~\cite{Davison-PAMI2007}  solutions, whereby the pose of an agent within an unknown map is tracked, have become a necessity with the advent of autonomous robots and Augmented Reality (AR) wearables that are equipped with cameras. The underlying geometry problem that needs solving is the Structure-from-Motion (SfM) problem that aims at recovering the structure of a scene, as well as the poses of a moving camera, from image correspondences \cite{Hartley-Book2004}.


In principle, visual data would suffice to solve SfM. In practice, however, apart from the scale ambiguity when a monocular sensor is used, the use of scene-dependent visual observation raises accuracy and efficiency issues. This led to the design of mixed sensors that combine visual sensing with other modalities. A successful paradigm is the fusion of visual with inertial data which has been proven to be beneficial for odometry solutions~\cite{Delmerico-ICRA2018}. The integration of inertial data, typically delivered by an Inertial Measurement Unit (IMU), not only provides valuable information  
for the ego-motion estimation, but it also resolves ambiguities of visual cues (low-texture, fast motion etc).

%

The resulting visual-inertial odometry (VIO) problem is usually cast into either a filtering formulation~\cite{Mourikis-ICRA2007,Li-ICRA2012} or a chain of optimizations~\cite{Leutenegger-IJRR2013,Indelman-RAS2013}. 
 Therefore, the initialization of the state  
 is required to either start or recover from divergence. The state typically includes the pose and the velocity of the sensor, while the reconstruction of the map points is implicitly required. When the sensor is strictly static, state initialization reduces into a simple orientation problem using only accelerometer data. However, when the system undergoes motion, the initialization becomes more difficult and visual-inertial SfM (vi-SfM \cite{Martinelli-IJCV2013}) must be solved. In addition, inexpensive inertial sensors and rolling-shutter cameras make vi-SfM even more challenging due to biased readings and sequential readout, respectively.
\begin{figure}[t]
  \begin{center}
  \scriptsize
      \psfrag{m}[c][c]{$\V{m}$}
      \psfrag{l1}[r][r]{$\lambda_1$}
      \psfrag{l2}{$\lambda_2$}
      \psfrag{u1}{$\V{u}_1$}
      \psfrag{u2}[r][r]{$\V{u}_2$}
      \psfrag{C1}[r][r]{$\MM{cam}_1$}
      \psfrag{I1}[r][r]{$\MM{imu}_1$}
      \psfrag{C2}[r][r]{$\MM{cam}_2$}
      \psfrag{I2}[l][l]{$\MM{imu}_2$}
      \psfrag{v}{\color[RGB]{255, 0, 0}$\V{v}_0$}
      \psfrag{g}[l][l]{\color[RGB]{255, 0, 0}$\V{g}_0$}
      \psfrag{tra}[l][l][1][-10]{trajectory}
      \psfrag{p}{$\V{p}_{I}$}
      
      \psfrag{mhat}[c][c]{$\hat{\V{m}}$}
      \psfrag{C1}[r][r]{$\MM{cam}_1$}
	 \psfrag{C3}{$\MM{cam}_3$}
         \psfrag{m1}[r][r]{$\hat{\V{m}}^{\{1\}}$}
	 \psfrag{m2}{$\hat{\V{m}}^{\{2\}}$}
	 \psfrag{m3}{$\hat{\V{m}}^{\{3\}}$}
         \psfrag{u1}[r][r]{$\V{u}_1$}
         \psfrag{u2}[r][r]{$\V{u}_2$}
         \psfrag{u3}[r][r]{$\V{u}_3$}
      \psfrag{I3}[l][l]{$\MM{imu}_3$}
            \psfrag{l3}{$\lambda_3$}
      \begin{tabular}{@{\hspace{2.5mm}}cc@{\hspace{-3.5mm}}c}
         \includegraphics[width=0.45\linewidth]{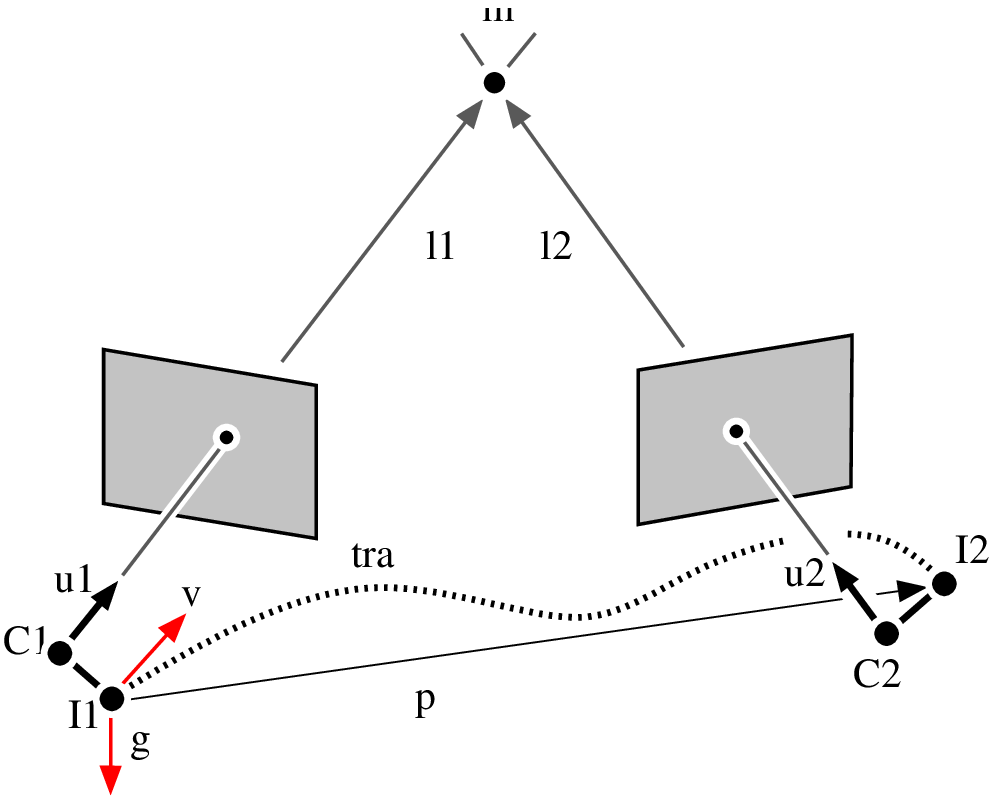} &~&
         \includegraphics[width=0.45\linewidth]{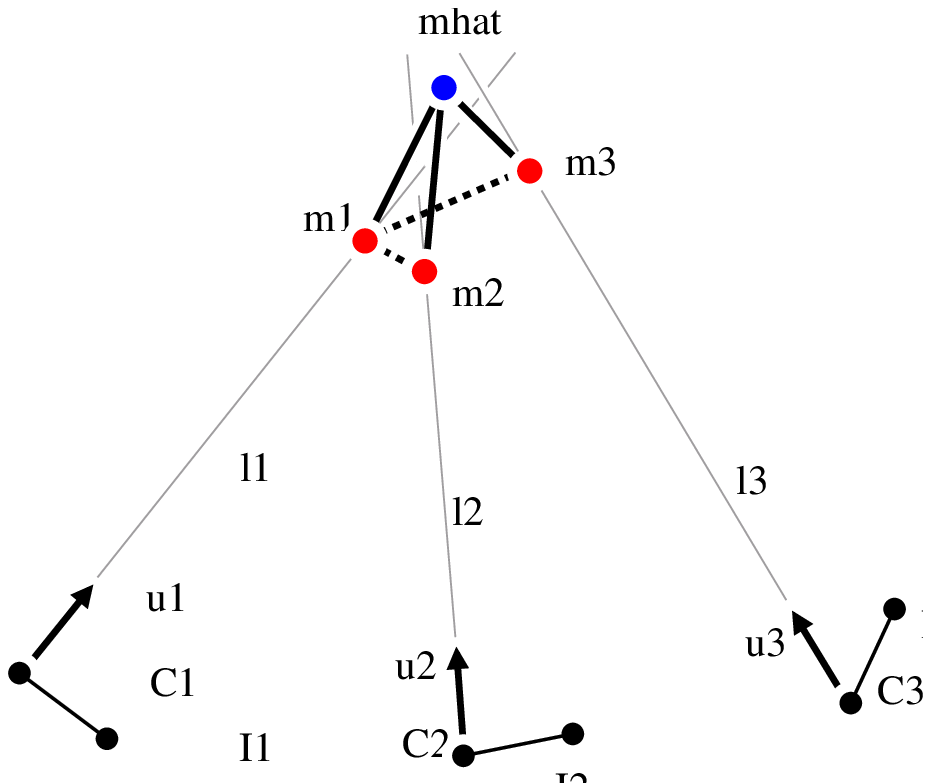} \\
         (a) &~& (b)\\[1ex]
     \end{tabular} 
    \end{center}
 \caption{(a) The \emph{visual-inertial triangulation} principle: the camera baseline is decomposed to the camera-to-IMU distances and to the IMU displacement $\V{p}_{I}$ that linearly depends on velocity $\V{v_0}$ and local gravity $\V{g}_0$ through the kinematic equation. (b) The multi-view case: The total distance between the single reconstruction $\hat{\V{m}}$ and all the candidates $\hat{\V{m}}^{\{i\}}$ (solid lines) is minimized by the proposed solver. Instead,~\cite{Martinelli-IJCV2013} minimizes the distance between the candidate pairs (dashed lines), that is, $\tilde{\V{m}}^{\{1\}}$ plays the role of $\hat{\V{m}}$. Only ideal conditions and perfect data make the two formulations equivalent.}
 \label{fig:3dDistanceError}
\end{figure}

Recently, \cite{Martinelli-IJCV2013} introduced a linear model for vi-SFM that builds on the triangulation principle. We refer here to this principle as \emph{visual-inertial triangulation} (see Fig.~\ref{fig:3dDistanceError}(a)). The derivation stems from the fact that the camera displacement can be expressed by a kinematic differential equation whereby, under some assumptions, unknown state and auxiliary parameters become linearly dependent. As a result, a closed-form solution for the problem in question becomes feasible. 
  
In this context, we build on the visual-inertial triangulation principle, but from the perspective of the multi-view midpoint algorithm~\cite{Sturm-CIV2006}. More specifically, instead of linking multiple pairs of image observations, we \emph{jointly} relate all the image observations with their generator, that is, the scene $3D$ point (see Fig.~\ref{fig:3dDistanceError}(b)). This leads to a different structure of the linear dependence among state and auxiliary parameters with two main advantages. Firstly, it allows the elimination of auxiliary variables at negligible cost, such that the initial velocity and orientation against the gravity axis can be determined by solving a $6\times 6$ linear system. Secondly, the joint dependence of all the point observations from the single yet unknown map point makes the estimator less biased. As a result, an inherently efficient and more accurate closed-form solution becomes available. The advantages against the formulation of \cite{Martinelli-IJCV2013} are discussed in detail in Sec. \ref{sec:closed-form-solution}.  As it is customary, we also combine the proposed solver with a non-linear refinement that better models any underlying non-linearity, such as the dependence on the gyroscope bias~\cite{Kaiser-RAL2017}.  The analytic Jacobians for the non-linear optimization are also provided.

To the best of our knowledge, this is the first work that focuses on the special structure of the resulting linear system. While most prior work solves a large linear system~\cite{Dong-IROS2012,Martinelli-IJCV2013,Kaiser-RAL2017,Mur-Artal-RAL2017}, we here show that this is unnecessary. The structure of the proposed system matrix allows for very cheap elimination, and hence, an efficient state initializer. The proposed elimination does not depend on any gravity related constraint that needs to be enforced~\cite{Martinelli-IJCV2013}. Rather, it separates the IMU state from the map points, that is, any constraint can be directly added to the eliminated system. 

\section{Related Work}\label{sec:RelatedWork}
While most of the methods assume known initial conditions for VIO~\cite{Mourikis-ICRA2007,Li-ICRA2012}, there has not been much related work that focuses on the initialization per se. 

Provided a calibrated device, the linear dependence of state parameters was discussed in~\cite{Martinelli-TRO2012}, whereby an observability analysis was presented and a closed-form initializer was derived. An extended work, though, that included a simpler closed-form and a thorough resolvability analysis for both biased and unbiased cases was then presented in~\cite{Martinelli-IJCV2013}. The latter constitutes the baseline for later work~\cite{Kaiser-RAL2017,Campos-ICRA2019}, as well as for our method.  

As far as the linear model is concerned, \cite{Martinelli-IJCV2013} relates corresponding visual observations through the camera baseline, which linearly depends on the unknown state parameters, that is, the velocity, the gravity in the IMU frame and the accelerometer bias. Each visual correspondence contributes three equations, while the distances between map points and the cameras become unknown parameters too. The resulting linear system is then solved, with an optional constraint on the gravity magnitude. The robustness of the method against biased IMU readings was investigated by~\cite{Kaiser-RAL2017} and, to account for the gyroscope bias, a non-linear refinement method was proposed. The work of~\cite{Campos-ICRA2019} then built on \cite{Martinelli-IJCV2013,Kaiser-RAL2017} and improved the method via multiple loops of visual-inertial bundle adjustments and consensus tests.

The above methods adopt an early fusion approach, a.k.a. tightly-coupled fusion. Instead, visual SfM problem can be first solved and IMU data can be later integrated in a more loosely-coupled framework~\cite{Kneip-IROS2011,Mur-Artal-RAL2017,Huang-TRO2020}. In this context,~\cite{Kneip-IROS2011} suggested using visual SfM to obtain camera velocity differences which are then combined with integrated IMU data to recover the scale and gravity direction. The initialization part of~\cite{Mur-Artal-RAL2017} used scaleless poses from ORB-SLAM~\cite{Mur-Artal-TRO2015} and then solved several sub-problems to initialize the state and the biases along with the absolute scale. This multi-step solution for the parameter initialization was then adapted in~\cite{Qin-IROS2011}.

The initialization problem becomes harder when the device is uncalibrated~\cite{Dong-IROS2012,Huang-TRO2020}. Even if the biases are known or ignored, the unknown orientation between camera and IMU makes the model non-linear and iterative optimization is necessary. In~\cite{Dong-IROS2012}, two solutions to estimate the unknown orientation are proposed, thus allowing solving a linear system that, in turn, initializes a non-linear estimator. 
Instead,~\cite{Huang-TRO2020} builds on the multi-step approach of~\cite{Mur-Artal-RAL2017} to jointly calibrate the extrinsics and initialize the state parameters. In a real scenario, however, the joint solution of calibration and initialization problem using only the very first few frames might make the pose tracking algorithm prone to diverge. 

All the above methods silently assume that visual observations come from a global-shutter sensor. Consumer devices, however, are mostly equipped with rolling-shutter cameras. This means that rolling-shutter effects need to be properly handled~\cite{Hedborg-CVPR2012,Ling-ECCV2018,Bapat-CVPR2018}. In the context of SLAM, rolling shutter  can be well modelled by continuous-time models that use temporal basis functions~\cite{Furgale-ICRA2012,Patron-Perez-IJCV2015,Ovren-CVPR2018}. These methods, however, do not focus on on the initialization problem, that is, their estimators are either partially initialized, e.g., from visual-only solvers, or even start from  identity poses and points at infinity. Since our test platform is a stereo rolling-shutter rig, we take into account the rolling-shutter readout time when implementing any method in Sec.~\ref{sec:experiments}.

\section{Proposed formulation}\label{sec:ProposedFormulation}

Assume a $3D$ point $\V{m}$ in a reference coordinate system (RCS) that is observed at $N$ different times by a moving camera via the transformation
\begin{equation} \label{eq:camera_projection}
 \lambda_i \M{R}_{C_i} \V{u}_i + \V{p}_{C_i} = \V{m}, ~~~i = 1,...,N,
\end{equation}
where $\V{u}_i$ is the normalized (calibrated) unit vector of the underlying image observation, $\lambda_i$ is the distance between the point and the camera, $\M{R}_{C_i}$\! is the matrix that characterizes the rotation from the camera coordinate system (CCS) to the RCS, $\V{p}_{C_i}$ is the camera position in the RCS, at the time $t_i = t(n_i T_s)$, $n_i \in \mathbb{N}$, and $T_s$ is a sufficiently low sampling time. Note that RCS is different than any CCS.

\begin{figure}[t]
  \begin{center}
         \includegraphics[width=0.69\linewidth]{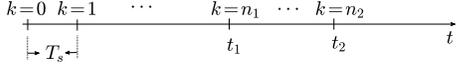}
 \caption{Sampling times: $T_s$ corresponds to the sampling period of inertial data; timestamps of visual observations, $t_1 = t(n_1T_s)$ and $t_2 = t(n_2T_s)$, coincide with irregular inertial sampling times.}
\label{fig:timeAxis}
\end{center}
\end{figure}

Assume also an intrinsically and extrinsically (against the camera) calibrated IMU that is rigidly mounted to the moving rig.\footnote{We silently assume that both IMU and camera are triggered by a common clock. In practice, a temporal calibrated offset aligns the time axes of the sensors.} Without loss of generality, the sampling period of the inertial signal can be set to $T_s$, as shown in Fig.~\ref{fig:timeAxis}. If we now consider the IMU frame at time $t_0=0$ as the RCS, the camera position $\V{p}_{C_i}$ can be written as
 \begin{equation} \label{eq:translation_decomposition}
\V{p}_{C_i} = \V{p}_{I_i}  +  \M{R}_{I_i}  \V{p}^I_C~\!,
\end{equation}
where $\M{R}_{I_i}$, $\V{p}_{I_i}$ are the orientation and position, respectively, of the IMU in the RCS at time $t_i$ and $ \V{p}^I_C$ is the known position of the camera in the IMU frame. 

Let us now assume a constant acceleration kinematic model \cite{Lupton-TRO2012} that describes the position of the IMU over time. Provided that  $\V{p}_{I_0} = \mathbf{0}$ and $\V{v}_0$ are the position and velocity, respectively, of the IMU in the RCS at time $t_0=0$, the successive integration of acceleration data results in the following equation,
 \begin{equation} \label{eq:kinematicEquation}
 \V{p}_{I_i}  = t_i\V{v}_{0} + \frac{t_i^2}{2}\M{R}_W\V{g}_{W}  + \frac{T_s^2}{2}\sum_{k=0}^{n_i-1}\beta_{ki} R_{I_k}(\boldsymbol{\alpha}_{I_k} + \V{b}_a)~\!,
\end{equation}
where $\V{g}_W$ is the gravity vector in the world coordinate system (WCS), $\M{R}_W$ is the matrix that represents the rotation from WCS to the RCS, $\boldsymbol{\alpha}_{I_k}$ is the measured acceleration at time $t_k$, $\V{b}_a$ is the accelerometer bias compensation that is considered constant for short integration times, and $\beta_{ki} = 2(n_i-k)-1$ is  the resulting coefficient from unfolding recursive integrations.

As mentioned, the IMU is internally calibrated and rigid corrections of gyroscope and accelerometer axes have been pre-applied. Sensor biases may be though affected by several sources and their online refinement is recommended. While the accelerometer bias offset $\V{b}_a$ is linearly added in \eqref{eq:kinematicEquation}, a gyroscope bias offset would break the linearity and its use through a non-linear refinement, when needed, is preferred \cite{Kaiser-RAL2017,Campos-ICRA2019}. Assuming now that the bias has been removed, any rotation matrix $R_{I_i}$ can be computed from integrating gyroscope data~\cite{Lupton-TRO2012},
 \begin{equation} \label{eq:gyroIntegration}
 \M{R}_{I_i}  = \prod_{k=0}^{n_i-1} \exp(\boldsymbol{\omega}_k T_s) = \exp(\boldsymbol{\omega}_0 T_s)\dots\exp(\boldsymbol{\omega}_{i-1} T_s)
\end{equation}
where $\boldsymbol{\omega}_k$ is the gyroscope measurement. As a result, $\M{R}_{C_i}$ can be as well estimated using the known orientation of the CCS in the IMU frame $\M{R}_{C}^{I}$, that is, $\M{R}_{C_i} = \M{R}_{I_i} {\M{R}_{C}^{I}}$.

The  equations \eqref{eq:camera_projection}, \eqref{eq:translation_decomposition} and \eqref{eq:kinematicEquation} can be combined into a single  matrix form as 
\begin{equation}\label{eq:model}
\left[t_i\M{I}_3~~~\frac{t_i^2}{2}\M{I}_3~~~\M{B}_i~~-\!\M{I}_3~~~\M{R}_{C_i} \V{u}_i  \right]
\left[\begin{array}{c}\V{v}_0 \\ \V{g}_0 \\ \V{b}_a \\ \V{m} \\ \lambda_i \end{array}\right] =  \V{c}_i~\!,
\end{equation}
where $\V{g}_0 = \M{R}_W\V{g}_{W}$ is the gravity in the RCS,~$\V{c}_i =  -\M{R}_{I_i}  \V{p}^C_I - \frac{T_s^2}{2}\sum_{k=0}^{n_i-1}\beta_{ki} R_{I_k}\boldsymbol{\alpha}_{I_k}$ is a constant vector that includes accumulation of weighted and rotated acceleration measurements, $\M{B}_i=\frac{T_s^2}{2}\sum_{k=0}^{n_i-1}\beta_{ki} R_{I_k}$ is a weighted sum of rotation matrices, and $\M{I}_3$ is the $3\times 3$ identity matrix.

Since $N$ observations of the point $\V{m}$ are available, one can easily extend  the above linear equations system. 
As a result, each visual observation adds three equations and one unknown $\lambda$ parameter, thus shaping a linear system of $3N\times(N+12)$ from a single point. Multiple points are typically needed and a large linear system is built.

Recall that the goal of the initialization is to estimate the initial velocity $\V{v}_0$ and the orientation $\M{R}_W$. It is customary to align the $z$-axis of the WCS with the gravity axis and set $\V{g}_W=[0,~0,~\gamma]^{\top}$, where $\gamma$ is the gravity magnitude. This makes $\V{g}_0$ a scaled version of the third column of matrix $\M{R}_W$, while $\|\V{g}_0\|_2 = \gamma$.  As a result, any rotation around the world gravity axis is not identifiable and $\M{R}_W$ is estimated up to this unknown (yaw) angle. Note also that $\V{b}_a$ is not separable from $\V{g}_0$ unless the system rotates, that is, $\M{R}_{I_k} \neq \M{I}_3$.\footnote{When $\M{R}_{I_k} = \M{I}_3$ then $\M{B}_i = \frac{t_i^2}{2}\M{I}_3$ which is equal to the coefficient of $\V{g}_0$.} As a result $\M{R}_{I_k} \neq \M{I}_3$ makes $\V{b}_a$ observable, while the constraint may be needed depending on the underlying case. E.g., in the particular case of rotation around at least two axes, the gravity constraint is not necessary for the biased case (see Property 15 in \cite{Martinelli-IJCV2013}).

\section{Closed-form solution}\label{sec:closed-form-solution}
Unlike \cite{Martinelli-IJCV2013}, we do not relate observation pairs. Instead, we add the unknown points, expressed in the RCS, into the parameter vector and directly relate every single point with its observations, that is, $\V{m}$ remains an unknown parameter of the linear system. Such an approach may initially result in an unknown vector of slightly higher dimension. However, as we see below, the matrix of the linear system has a simpler form and any elimination can be obtained at no cost, that is, without any matrix inversion or decomposition. Moreover, the direct reconstruction of the points in the RCS comes as a by-product. 

Let us consider $M$ map points, stacked into a vector $\bm{\mu} = [\V{m}_1^\top, ..., \V{m}_M^\top]^\top$,and let $\lambda_{ji}$ and $\V{u}_{ji}$ denote the corresponding rays and distances, respectively. For the sake of simplicity, we assume that each point has the same number of $N$ observations (captured at $N$ different times) while in practice each point can have a different number of observations. If we set $\bm{\lambda} = [\lambda_{11}...,\lambda_{MN}]^\top$ and $\V{z} = [\V{v}_0^\top, \V{g}_0^\top, \V{b}_a^\top]^\top$, the entire linear system can be written as
\begin{equation}\label{eq:totalLinearSystem}
\left[\M{V} ~|~ \M{W} ~|~ \M{Q}\right] \left[\begin{array}{c}\V{z} \\ \bm{\mu} \\ \bm{\lambda} \end{array}\right] =  \V{c}~\!,
\end{equation}
where  $ \M{V}$ is a $3MN\times 9$ matrix, $ \M{W}$ is a $3MN\times 3M$ block matrix with diagonal structure, $\M{Q}$ is a $3MN\times MN$ block matrix with diagonal structure and $\V{c}$ is a constant vector of length $3MN$:
\begin{equation}\label{eq:Vmatrix}
 \M{V} =
\left[
\begin{array}{ccc}
    t_{11}\M{I}_3 &\frac{t_{11}^2}{2}\M{I}_3 &\M{B}_{11} \\
    \vdots & \vdots & \vdots \\
    t_{MN}\M{I}_3 &\frac{t_{MN}^2}{2}\M{I}_3 &\M{B}_{MN} \\
    \end{array}
  \right]~\!,
\end{equation}

\begin{equation}\label{eq:Wmatrix}
 \M{W} =
\left[
\begin{array}{ccc}
    \M{Y}_{1} & & \\
    & \ddots & \\
    & & \M{Y}_{M}
    \end{array}
  \right]~\!,
\end{equation}

\begin{equation}\label{eq:Umatrix}
 \M{Q} =
\left[
\begin{array}{ccc}
    \V{q}_{11} & & \\
    & \ddots & \\
    & & \V{q}_{MN}
    \end{array}
  \right]~\!,
\end{equation}

\begin{equation}\label{eq:constantVector}
 \V{c} =
\left[
\begin{array}{c}
    \V{c}_{11}\\
    \vdots \\
    \V{c}_{MN}\\
    \end{array}
  \right]~\!,
\end{equation}
with $\M{Y}_j = -[\M{I}_3, \hdots, \M{I}_3]^{\top}$ being a $3N\times 3$ block and $\V{q}_{ji} = \M{R}_{C_{ji}}\V{u}_{ji}$. 

The vector $\bm{\lambda}$ contains auxiliary variables and its elimination is meaningful. Commonly, one would multiply from the left with the projection matrix $\M{P} = \M{I} -  \M{Q}(\M{Q}^{\top}\M{Q})^{-1}\M{Q}^{\top}$.
Recall, however, that each block of $\M{Q}$ is a unit vector, hence $(\M{Q}^{\top}\M{Q})^{-1} = \M{I}$. As a consequence, the block diagonal matrix $\M{P} = \M{I} -  \M{Q}\M{Q}^{\top}$ can be computed without any inversion and such an elimination comes at negligible cost. The system one needs to initially construct is the following:
\begin{equation}\label{eq:totalLinearSystemEliminated}
\left[\M{P}\M{V} ~|~ \M{P}\M{W}\right] \left[\begin{array}{c}\V{z} \\ \bm{\mu}\end{array}\right] =  \M{P}\V{c}.
\end{equation}

It now becomes evident that the linear system is smaller than the one of \cite{Martinelli-IJCV2013,Kaiser-RAL2017} since $M \ll MN$. Note that homogeneous equations that relate pairs of $\lambda$-based reconstructed points are added in the linear system of \cite{Martinelli-IJCV2013}, thereby increasing the number of rows. We do not add such constraints here since all the image observations of a single point are \emph{jointly} related through a single unknown parameter. 

We now proceed with a second elimination step that further reduces the above linear system into one that only solves for the IMU state. One can optionally back-substitute to compute the points, when needed. To this end, we apply the projection operator $\M{I - PWHW^{\top}P^{\top}}$, where $\M{H = (W^{\top}PW)^{-1}}$ since $\M{P}$ is symmetric and idempotent. However, it is straightforward to show that $\M{W^{\top}PW}$ is a block diagonal matrix of size $3M\times 3M$, with each block being defined by $N(\M{I}_3 -\frac{1}{N} \sum_{i=1}^N \V{q}_{ji}\V{q}_{ji}^{\top})$. Hence, the computation of $\M{H}$ requires inverting each $3\times 3$ block, which is given by a simple analytical formula. Alternatively, one could make use of the Sherman-Morrison formula \cite{Golub-Matrix2013} for an inversion-free recursive computation with rank-1 updates. 

The elimination of map points finally leads to the following minimal system
\begin{equation}\label{eq:totalLinearSystemEliminatedTwice}
\M{P}\M{G}\M{V} \V{z}=  \M{PG}\V{c}~,
\end{equation}
where $\M{P=I-QQ^{\top}}$ and $\M{G = I - WHW^{\top}P}$. Apart from the fact that $\M{P}$ is a block diagonal matrix, the computation of $\M{WHW^{\top}}$ from $\M{H}$ involves only additions since all blocks of $\M{W}$ are identity matrices. As a consequence, we end up with a $9\times 9$, or a $6\times 6$ in the unbiased case, linear system that can be very efficiently built. Still, the norm equality constraint $\|\V{g}_0\|_2 = \gamma$ can be optionally added. There are several options to solve the resulting constrained problem, e.g., solving the unconstrained linear system followed by a one-step refiner, adding a quadratic constraint in a convex optimization framework, or applying QR decomposition to name a few. Additionally, any weighting scheme per observation or per map point easily applies.


%

When the point reconstruction is required, one can use the following equation to compute the coordinates:
\begin{equation}\label{eq:mapPointsSolution}
\bm{\mu}=
\M{H}\M{W}^{\top}\M{P}(\V{c} - \M{V} \V{z}^*) ,
\end{equation}
where $\V{z}^*$ is the solution of \eqref{eq:totalLinearSystemEliminatedTwice}.

Finally, the rotation matrix $\M{R}_W$ is computed from the angle between the vectors $\V{g}_0$ and $\V{g}_W$, while $\V{v}_0$ and $\V{m}_j$ are expressed in the WCS by $\M{R}_W^{\top}\V{v}_0$ and $\M{R}_W^{\top}\V{m}_j$, respectively. Since the origin of the WCS can be arbitrarily chosen, it can be identified with the origin of the RCS. 


The above formulation can be seen as a generalization of the multi-view midpoint triangulation algorithm~\cite{Sturm-CIV2006}, which builds on known poses to reconstruct the point that is closest (on average) to the observation rays. Here,  the poses are unknown. The camera positions depend on the same unknown set, which makes the reconstruction of different points dependent on each other. One might assume the parallelism of $\V{m} -\V{p}_{C_i}$ with $\V{u}_i$, thus setting their cross product equal to zero and ignoring $\lambda_i$ (DLT,~\cite{Hartley-Book2004}). However, the midpoint algorithm is simpler and more efficient for the multi-view case, in particular here where each point is not reconstructed independently. Moreover, it directly provides the sign of $\lambda$'s for a cheirality check.

\subsection{Rolling shutter} Unlike global shutter cameras that have single exposure-then-readout step for the whole image, rolling shutter (RS) cameras have a multi-step mechanism that captures the image rows sequentially, at different times. We deliberately refer to the time index in \eqref{eq:camera_projection} to consolidate these two cases.  Simply, all the visual observations of a single image have the same timestamp in global shutter mode. Instead, the timestamp of a visual observation of a rolling-shutter image can be given by $t_i = \tau_i + \tilde{u}_i^y\Delta\tau$, where $\tau_i$ is the timestamp of the first image row, $\tilde{u}_i^y$ is the lens-distorted y-coordinate (image row) of the projection of $\V{m}$ on the image and $
\Delta\tau$ is the readout time per image row.  As a result, each row has a different pose and~\eqref{eq:kinematicEquation} differs per image feature. More than one IMU samples correspond to a single image and an interpolation scheme can provide an IMU sample per $t_i$.

\subsection{Stereo camera} 

The proposed modelling is valid with either a monocular or a binocular sensor. Recall that $\V{m}$ is expressed in RCS and $\V{u}_i$ may regard any frame of either sensor. The integration time needed to reliably initialize the state may be different though. The stereo baseline leads to larger camera displacement, which in turn leads to better triangulation. For instance, given two successive stereo frames, the displacement from the \emph{current left} to the \emph{next left} camera is most of the times smaller than the distance between the \emph{current left} and the \emph{next right} camera. As a result, the integration time that is needed to cover a sufficient baseline is smaller, that is, less frames can be considered.

\subsection{Resolvability}
The resolvability of vi-SfM problem is discussed in detail in~\cite{Martinelli-IJCV2013}. Our solver differentiates in the way Eq.~\eqref{eq:model} is used, that is, the linear dependencies remain the same. Provided a varying acceleration, a minimum number of $5$ frames would suffice for a unique solution, even with a single point. When the system also rotates in 3D (around two or more axes), the biased case is uniquely solvable, when at least $6$ frames are used. The use of a second or third point relaxes these constraints in some cases~\cite{Martinelli-IJCV2013}. In practice, the use of a bunch of long tracks is recommended. Therefore, the above numbers could regard non-successive frames.

The stereo camera makes the problem solvable with less frames, since points are observable even from single stereo frames. For instance, when acceleration and rotation vary, $3$ frames would suffice to estimate the state in the unbiased case for any number of points, while one more frame is required when bias is included. In the particular case of  RS cameras, even less frames make the problem solvable because each scanline can be seen as a different "frame". The analysis of several motion and structure cases~\cite{Martinelli-IJCV2013} for stereo and/or rolling shutter cameras is long and we leave it for a feature work. It is worth mentioning that, in practice, more frames are required for a reliable solution.

\subsection{Outlier handling} So far, we silently assume that the visual correspondences are inliers, up to a reasonable tracking error. In practice, the tracks may include outliers. The above solution can be used as a minimal solver combined with a RANSAC-like scheme to cope with the outliers. But this would make the initializer quite slow. A better approach is to combine RANSAC with the tracker that provides visual correspondences, so that the solver receives outlier-free data. As an example, RANSAC on fundamental matrix removes inconsistent matches in~\cite{Campos-ICRA2019}. In Sec.\ref{sec:experiments}, we adopt the same selection scheme on raw matches that come from the ECC tracker~\cite{Evangelidis-PAMI2008} on FAST corners~\cite{Rosten-PAMI2008}. Although the RS effect makes the essential matrix globally invalid, it is sufficient to detect outliers given that any time-varying rotation is compensated via gyroscope data integration. Instead, the generalized essential matrix~\cite{Dai-CVPR2016} can be used when the RS effect is quite strong (very fast camera motion).

\subsection{Relation to \cite{Martinelli-IJCV2013}}
Eq.~\eqref{eq:model} is also used as the starting equation in~\cite{Martinelli-IJCV2013}, but pairwise ray differences eliminate the unknown point. However, all the possible pairs should be considered for an equivalent solution, since \eqref{eq:model} does not exactly hold. In addition, the mutually dependent reconstruction of multiple points makes the two solutions even more different. 
The advantages of the proposed solver compared to \cite{Martinelli-IJCV2013,Kaiser-RAL2017} can be summarized as follows:
\begin{itemize}
\item The linear system has an inherently simpler structure and auxiliary parameters are eliminated at negligible cost. This leads to a more efficient solution that only requires inverting or decomposing a very small matrix.
The elimination of $\lambda$'s in \cite{Martinelli-IJCV2013,Kaiser-RAL2017} would require inverting or decomposing a large sparse matrix with more complicated structure.
\item The proposed formulation leads to a linear system with uniquely defined structure. In contrast, the structure of  the linear system of \cite{Martinelli-IJCV2013} depends on how the observations pairs are combined and on how the points appear in frames. Note that, in practice, each point appears at different frames.
\item The reconstruction of the map points in a single RCS is directly obtained by the linear solver. When requested, it is the linear solver that directly estimates their coordinates. In \cite{Martinelli-IJCV2013,Kaiser-RAL2017}, one would typically average the many putative reconstructions per point, or choose the reconstruction in one of the CCS,  while different points may be reconstructed in different CCS (partial tracks).
\item The estimation is better conditioned since all the point observations are jointly and symmetrically related through the single yet unknown point that generates them.
\item The model naturally extends to a bundle adjustment scheme with the same parameters, e.g. by applying a projection operator (Sec.~\ref{sec:nonLinearRefinement}). Instead, initial map points in a single RCS or CCS should be pre-computed when the solver of~\cite{Martinelli-IJCV2013,Kaiser-RAL2017} is used.
\end{itemize}



\section{Non-linear refinement}\label{sec:nonLinearRefinement}
The underlying application may require high accuracy while the available hardware may support computationally demanding operations. Therefore, we suggest a further refinement of the IMU state and the reconstructed points in an iterative optimization framework. We do not solve a multi-keyframe Visual-Inertial Bundle Adjustment problem whereby multiple states  are optimized~\cite{Campos-ICRA2019}. Instead, we simply optimize the image reprojection error \wrt  the initial single-frame state and structure, and optionally the biases.

As seen in Fig.~\ref{fig:3dDistanceError} , the solver in \eqref{eq:totalLinearSystemEliminatedTwice} minimizes the average $3D$. Despite the geometric nature, visual observations are back-projected in $3D$ space through an unknown depth, which may give some unwanted freedom to the solver. Therefore, the projection of the error distance onto a manifold (surface) that is directly observed makes more sense,  where one of the two vectors remains constant is more meaningful, while is makes $\lambda$ disappear. Commonly, the image itself or the calibrated image plane at $z=1$ of the CCS is used. 

Let us denote the total error that needs minimizing as
\begin{equation}\label{eq:totalError}
f(\V{x}) = \sum_{i,j} d(\pi(\M{R}_{C_i}^{\top}\V{m}_j - \M{R}_{C_i}^{\top}\V{p}_{C_i}), \pi(\V{u}_{ji}) )
\end{equation}
where $\V{x} = [\V{v_0}^{{\top}}, \V{g}_0^{\top}, \V{b}_a^{\top}, \V{m}_1^{\top}, \dots \V{m}_M^{\top}]^{\top}$, $d(\cdot, \cdot)$ is the (squared) Euclidean distance of the arguments, and $\pi(\V{u}) = [u_x/u_z,~u_y/u_z]^\top$ is the common perspective projection. 



When the constraint $\|\V{g}_0\|_2=\gamma$ must be enforced, a constrained optimization can be avoided by a proper parameterization of $\V{g}_0$. Since the rotation around the gravity axis is not observable, the unknown rotation can be paremeterized by the axis-angle vector $\boldsymbol{\phi}=[\phi_x, \phi_y, 0]^{\top}$. Using the Rodrigues formula~\cite{Forster-TRO2017} (exponential map) that provides $\M{R}_W$ from $\boldsymbol{\phi}$, the equation $\V{g}_0=\M{R}_W\V{g}_W$ leads to the parameterization
\begin{equation}\label{eq:gravityModel}
\V{g}_0\! =\! \gamma \left[   
\begin{array}{ccc}
\frac{\sin(\|\V{\boldsymbol{\phi}\|)}}{\|\boldsymbol{\phi}\|}\phi_y, & \! 
-\frac{\sin(\|\V{\boldsymbol{\phi}\|)}}{\|\boldsymbol{\phi}\|}\phi_x, & \!
\cos(\|\boldsymbol{\phi}\|)
\end{array}
\right]^\top,
\end{equation}
which makes the constraint valid.

The gyroscope bias can be also inserted into the model in a non-linear way \cite{Forster-TRO2017}. In such a case,  the unknown vector $\V{x}$ is augmented by an extra parameter $\V{b}_g$. 
We model the gyroscope bias in the experimental section to evaluate its contribution into the parameter estimation.

The Jacobians of the linearized form of $f(\V{x})$ with respect to the parameters (including the gyroscope bias) are given in the Appendix \ref{appendix}.

\section{Experiments}\label{sec:experiments}

\subsection{Experimental setup}

We are interested in experimenting with a stereo rolling shutter (RS) camera rigged with an IMU. In order to get realistic data with ground truth (GT) structure and poses, we process data from Snap Spectacles. States resulting from a Kalman filter on visual-inertial data play the role of GT states and high-order splines on IMU data provide ideal gyroscope and acceleration readings, such that a continuous integrator perfectly interpolates between the filter states. IMU data are then sampled at $800$Hz and finally, noise and time varying biases are added based on the calibrated variances of the used device. The frame readout time is $10$ms, that is, $8$ IMU samples are available per frame.

\begin{figure*}[t]
  \begin{center}
      \begin{tabular}{@{\hspace{0mm}}c@{\hspace{2mm}}c@{\hspace{2mm}}c}
               		\includegraphics[width=0.33\linewidth]{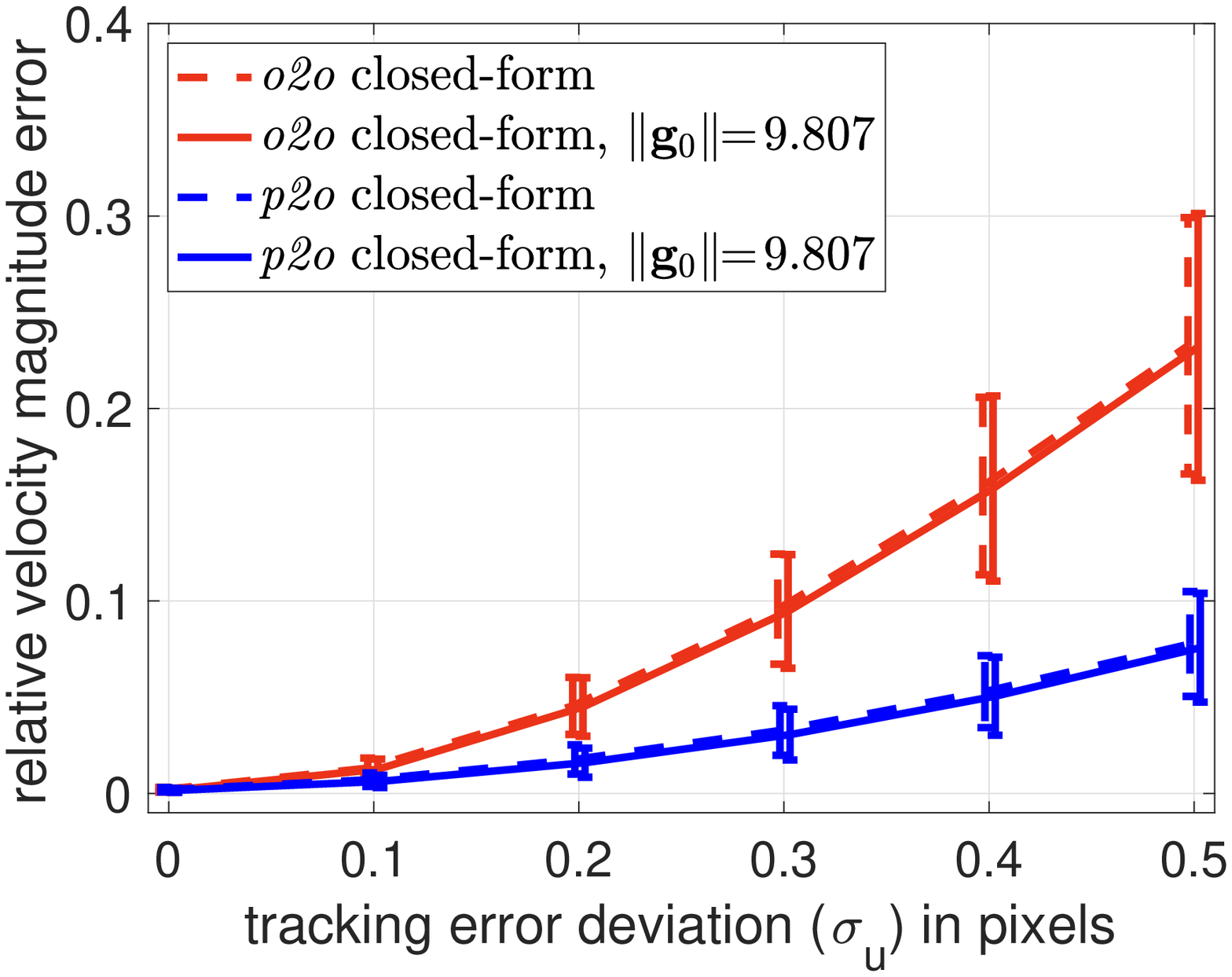} &
         \includegraphics[width=0.33\linewidth]{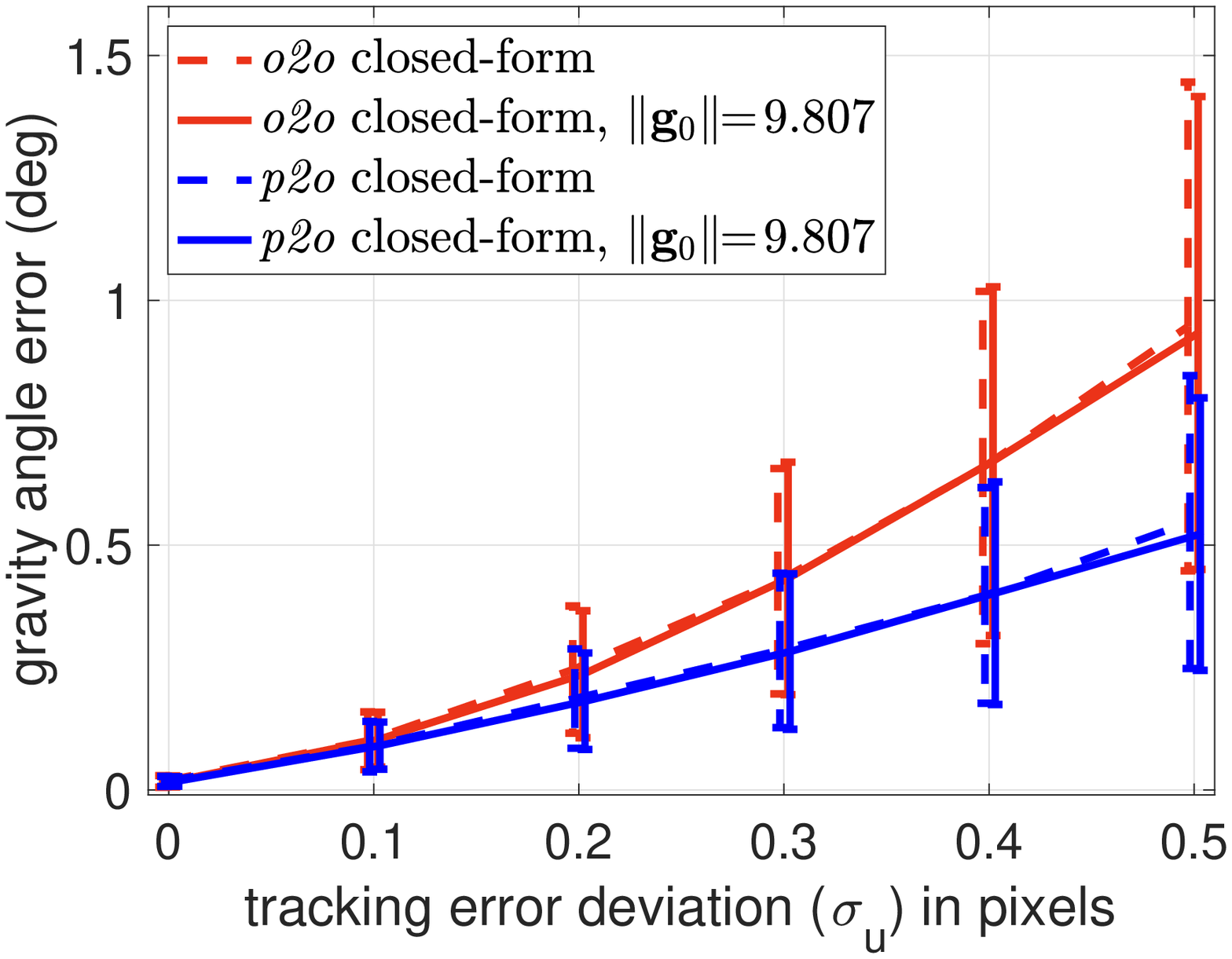} &
         		\includegraphics[width=0.33\linewidth]{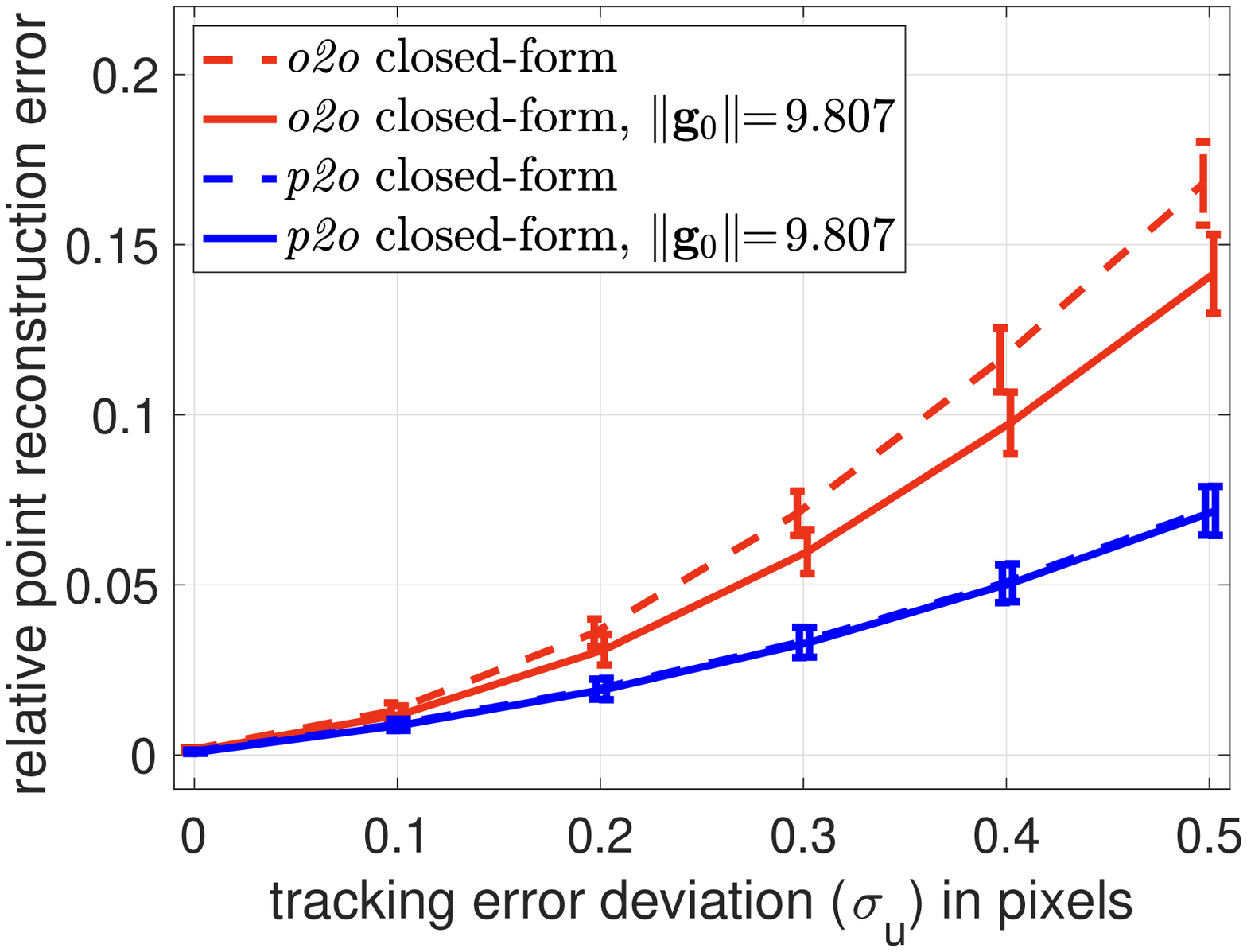} \\
         (a) & (b) & (c)\\[1ex]
     \end{tabular} 
    \end{center} \vspace{-0.3cm}
 \caption{(a) Velocity estimation, (b) gravity orientation estimation and (c) point reconstruction error as a function of point tracking error; the integration time is $0.46$ seconds ($N_f=3$).}
 \label{fig:performanceWrtTrackingError}
\end{figure*}
\begin{figure*}[t]
  \begin{center}
      \begin{tabular}{@{\hspace{0mm}}c@{\hspace{3.3mm}}c@{\hspace{3mm}}c}
               		\includegraphics[width=0.33\linewidth]{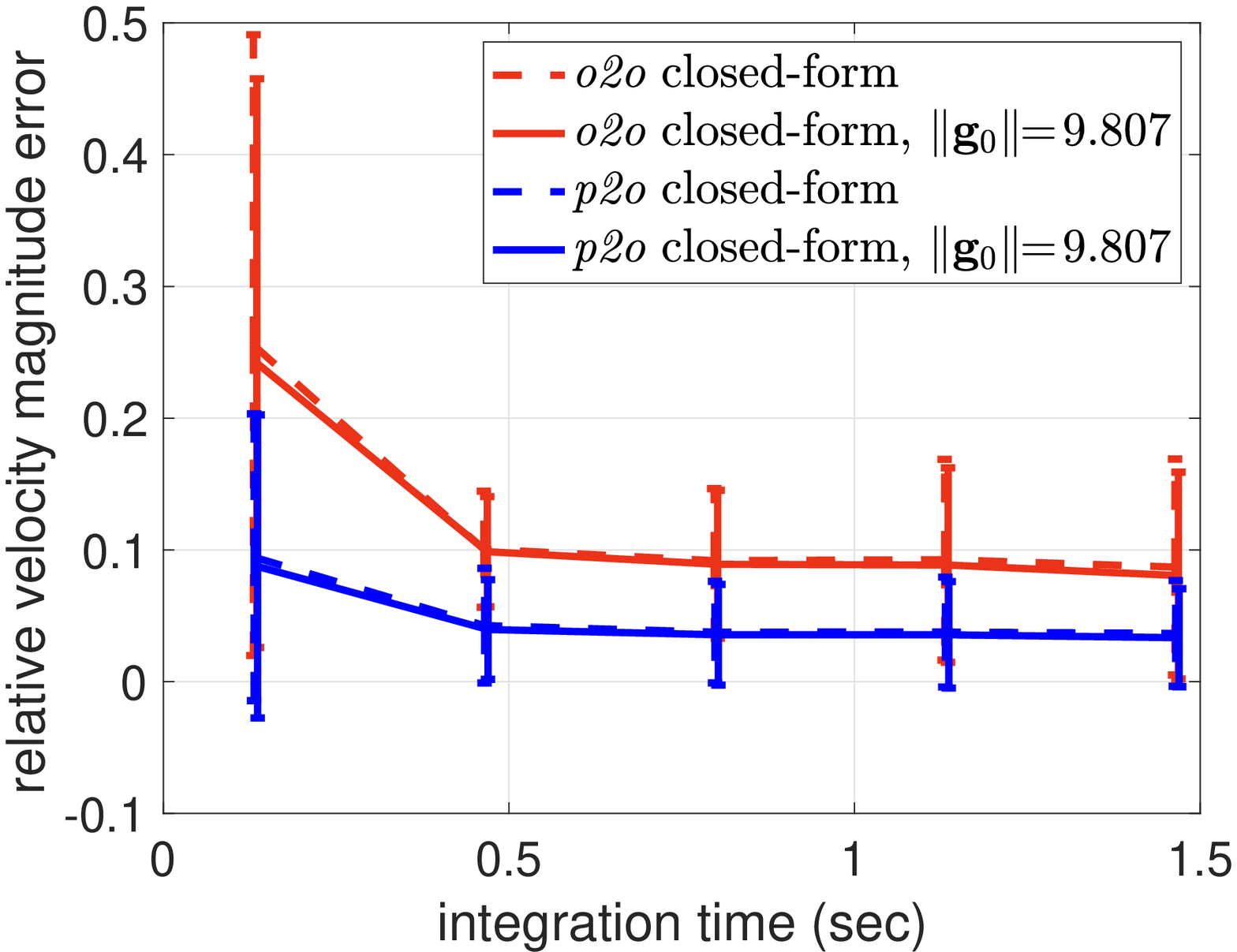} &
         \includegraphics[width=0.31\linewidth]{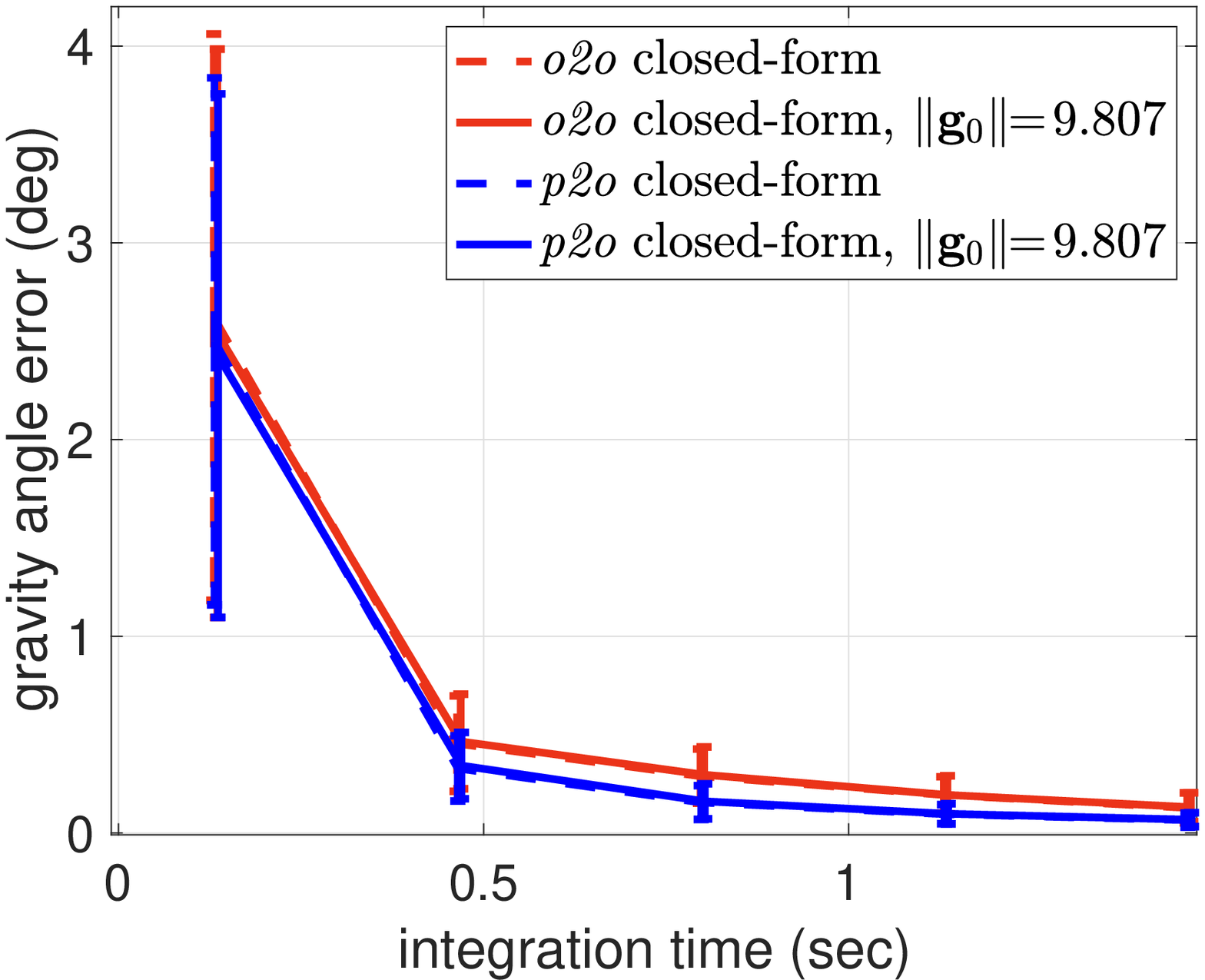} &
         		\includegraphics[width=0.33\linewidth]{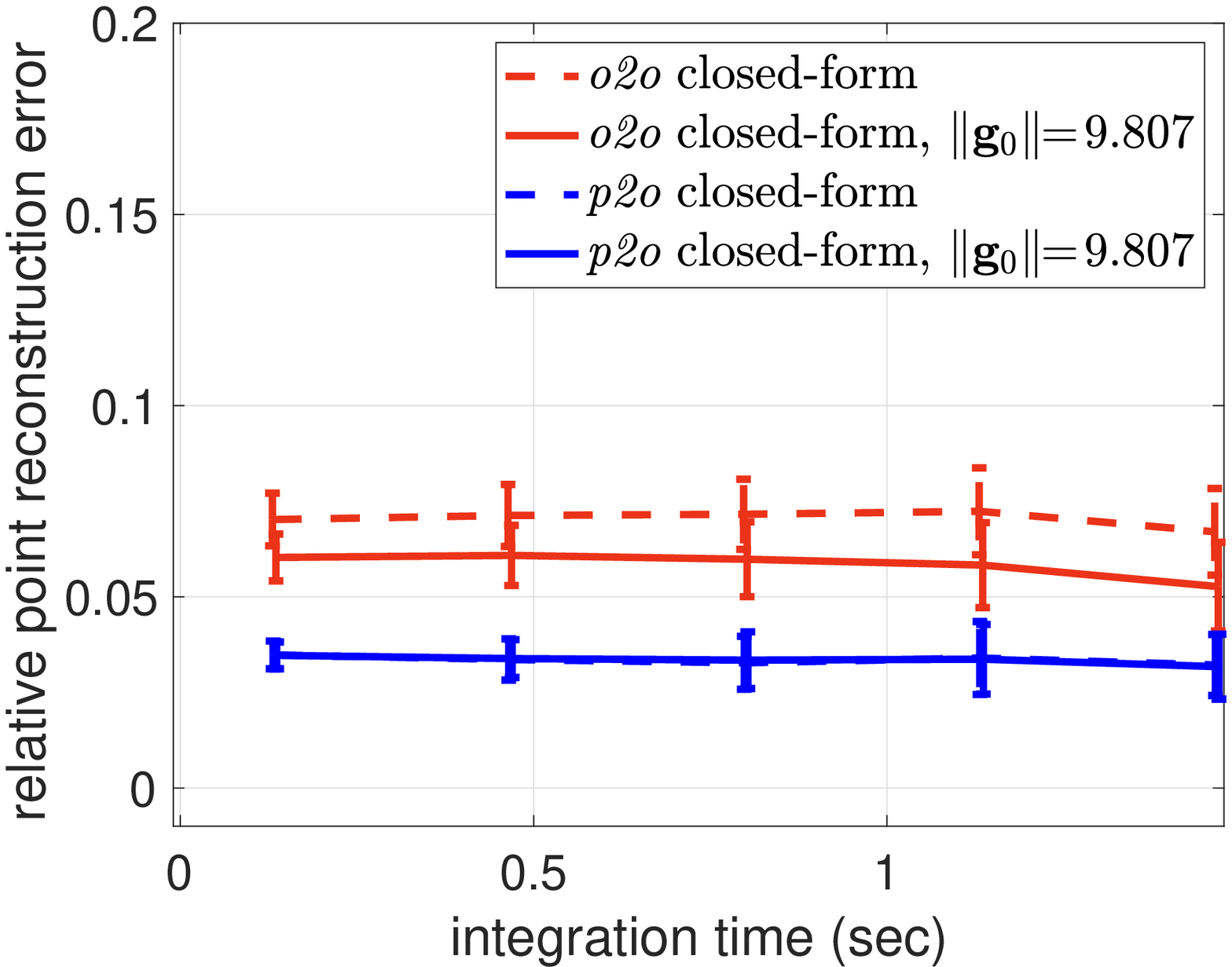} \\
         (a) & (b) & (c)\\[1ex]
     \end{tabular} 
    \end{center}\vspace{-0.3cm}
 \caption{(a) Velocity estimation, (b) gravity orientation estimation and (c) point reconstruction error as function of integration time; the point tracking error deviation is $0.3$ pixels.}
 \label{fig:performanceWrtIntegrationTime}
\end{figure*}

A virtual stereo rolling shutter camera of VGA resolution follows the resulting trajectory within a virtual $3D$ scene and images are rendered at 30Hz.\footnote{Unreal Engine is used~\cite{Unreal}.} When GT image correspondences are needed, single virtual $3D$ points along with their reprojections are created. Given a reference image, we back-project $100$ evenly spaced image points with random depth in range [$1$m,~$15$m] and the points are in turn re-projected into adjacent frames. As mentioned, we get real correspondences from rendered images using an ECC-based tracker~\cite{Evangelidis-PAMI2008} on FAST corners~\cite{Rosten-PAMI2008}, while a RANSAC-based scheme removes outliers. The tracked features do not necessarily appear in any frame of the time window. We consider tracks from $5$ or $7$ stereo frames, but we modify the frame downsamping factor to change the integration time and the rig displacement. A linear interpolation scheme estimates IMU samples at feature times. 

In all the experiments below, the initialization problem is solved from scratch per tested frame window, without using any prior information from previous solutions.



\subsection{Closed-form performance evaluation} 
We compare the performance of the proposed solver against the solver of \cite{Martinelli-IJCV2013}. 
We do not consider biases here and we deal with them below when non-linear refinement is employed. We refer to the proposed solver as \emph{point-to-observation} (\emph{p2o}) pairing scheme as opposed to \emph{observation-to-observation} (\emph{o2o}) pairing paradigm of  \cite{Martinelli-IJCV2013,Kaiser-RAL2017}.

The GT states regard a sequence from a Spectacles wearer who is almost static for about $1$ second and s/he then walks forward for $12$ seconds while looking around. Such a sequence mixes translational and rotational movements while it includes instant stationary parts. We here use virtual points and the tracks of GT image observations that are affected by additive Gaussian noise of known deviation $\sigma_u$. 

First, we test the robustness of the solvers in terms of the tracking error, which found to be the dominant parameter that affects the performance. For each value of $\sigma_u$ in the range $[0, 0.5]$ pixel, a sliding window of $5$ temporarily downsampled frames is used. The downsampling factor is $N_f =3$, thus defining a total integration time of $0.46$s. Any frame window with GT velocity magnitude below $0.01$m/s is discarded. In total, $50$ realizations per window are executed. 
The \emph{relative} magnitude error and the angular error are the evaluation criteria used to quantify the velocity and gravity  direction estimation, respectively. As for the point reconstruction error, the error distance per point is normalized by its depth. 
The average error as a function of $\sigma_u$, with and without the gravity norm constraint, is shown in Fig.~\ref{fig:performanceWrtTrackingError}. As seen, the proposed \emph{p2o} formulation is more robust and provides more accurate estimations, while its superiority against \emph{o2o} formulation grows with the tracking error. 
When the gravity norm constraint is enforced, the performance improvement is not noticeable in most of the cases.

\begin{figure*}[t]
  \begin{center}
      \begin{tabular}{@{\hspace{0mm}}c@{\hspace{2mm}}c@{\hspace{2mm}}c}
         \includegraphics[width=0.33\linewidth]{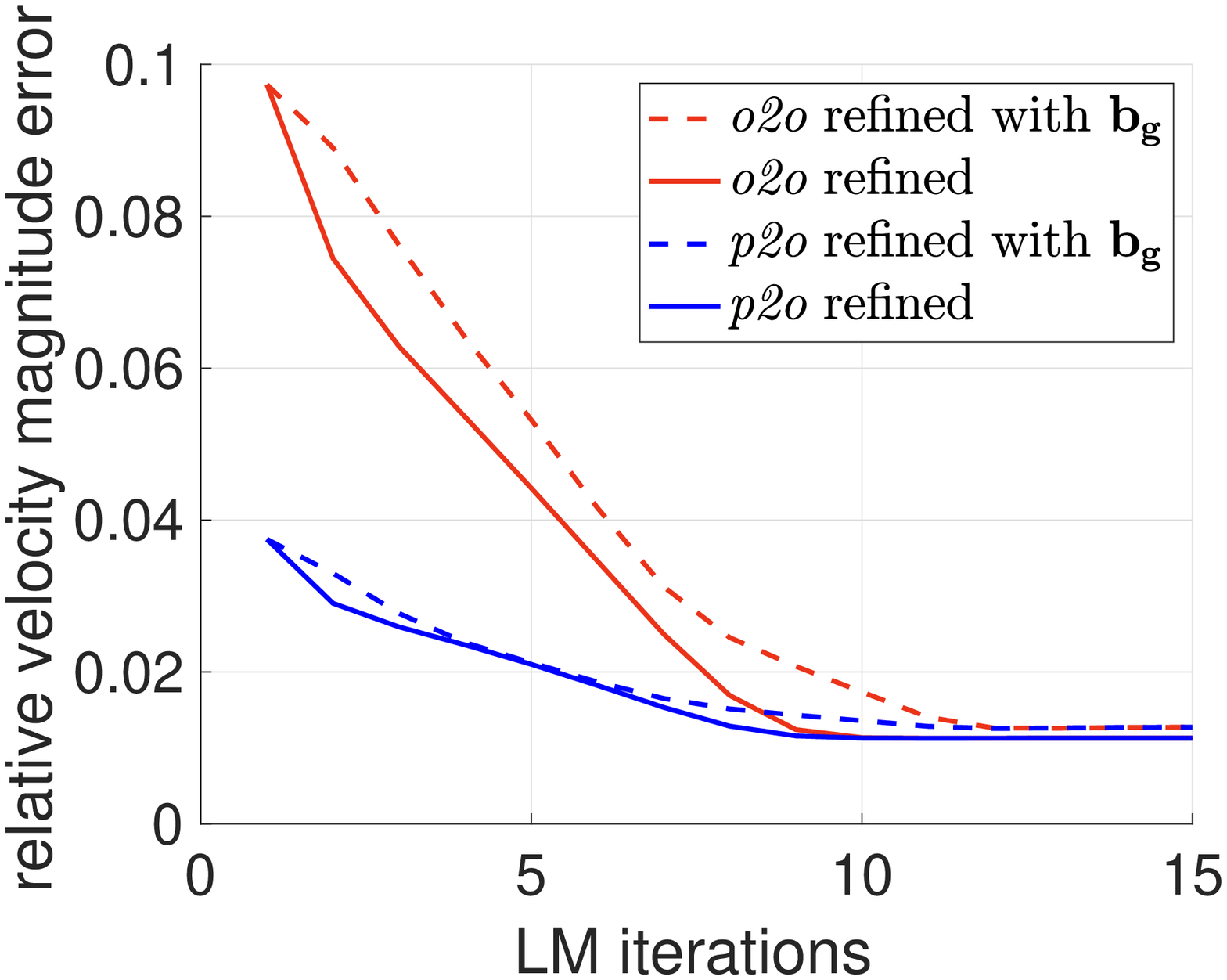} &
         \includegraphics[width=0.33\linewidth]{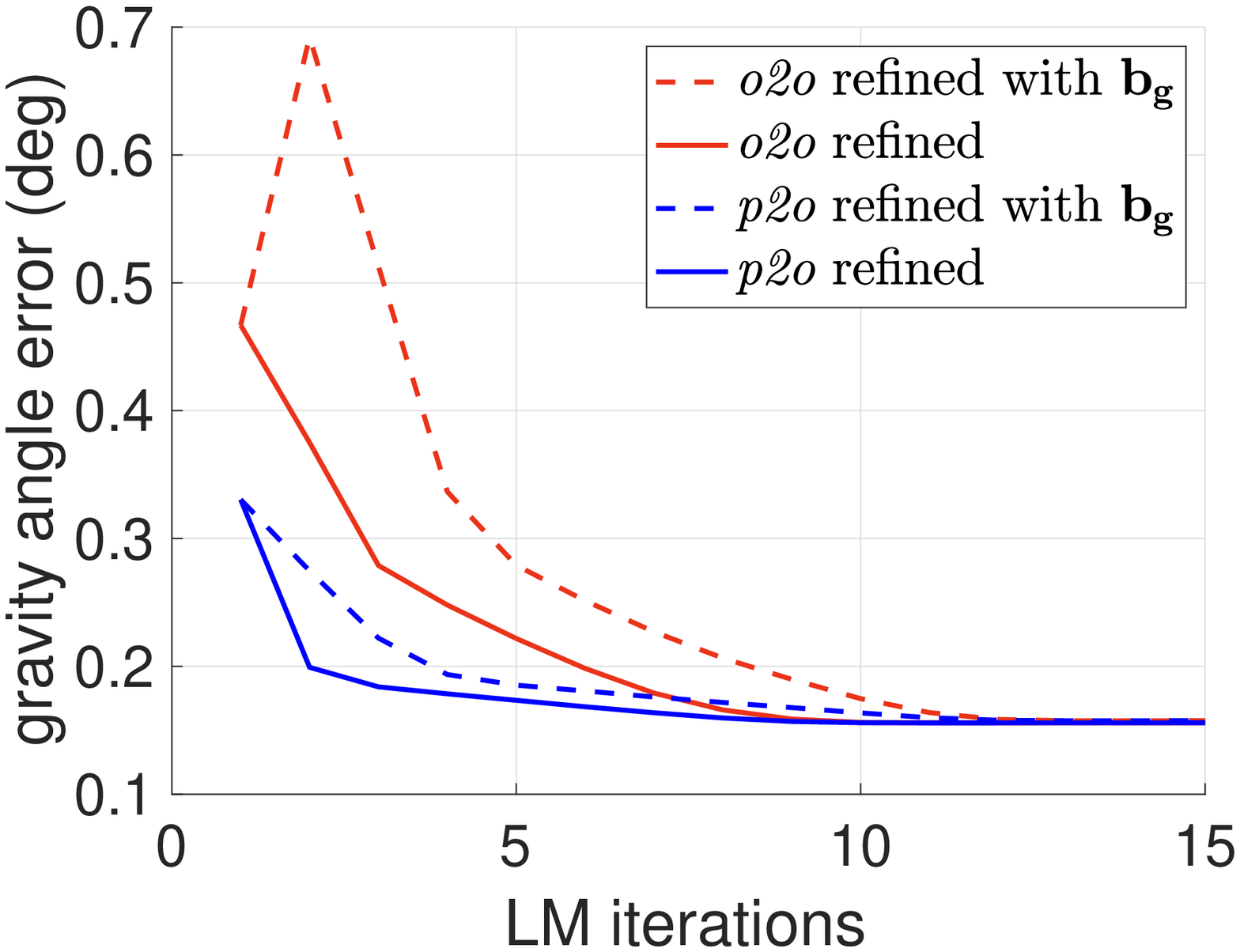} &
        \includegraphics[width=0.33\linewidth]{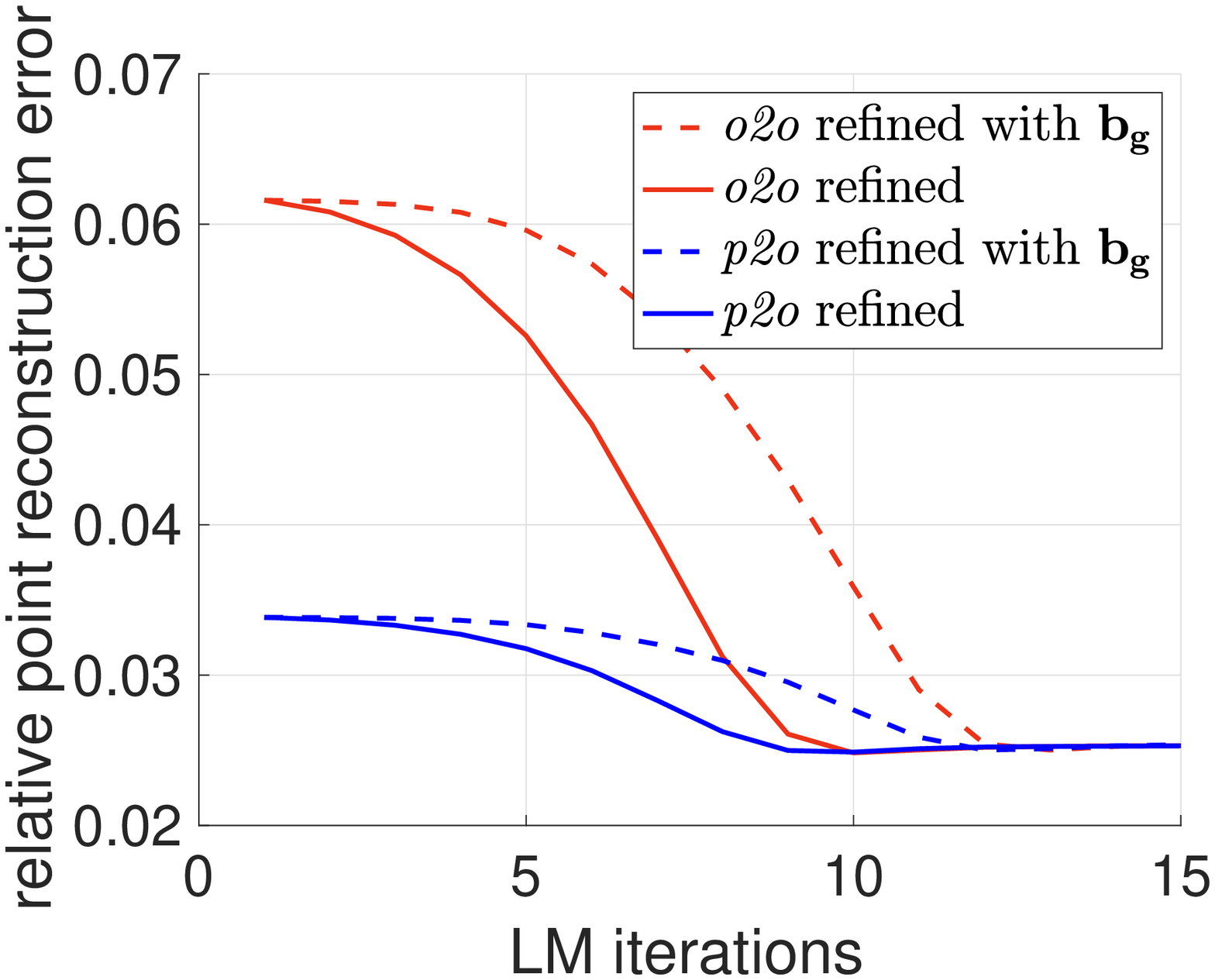} \\
         (a) & (b) & (c)\\[1ex]
     \end{tabular} 
    \end{center}
 \caption{(a) Velocity estimation, (b) gravity orientation estimation and (c) point reconstruction error attained per iteration with non-linear refinement; the point tracking error deviation is $0.3$ pixels and the integration time is $0.46$ seconds.}
 \label{fig:nonLinearRefinementPerformance}
\end{figure*}
\begin{figure*}[t]
  \begin{center}
      \begin{tabular}{@{\hspace{0mm}}c@{\hspace{2mm}}c@{\hspace{2mm}}c}
         \includegraphics[width=0.33\linewidth]{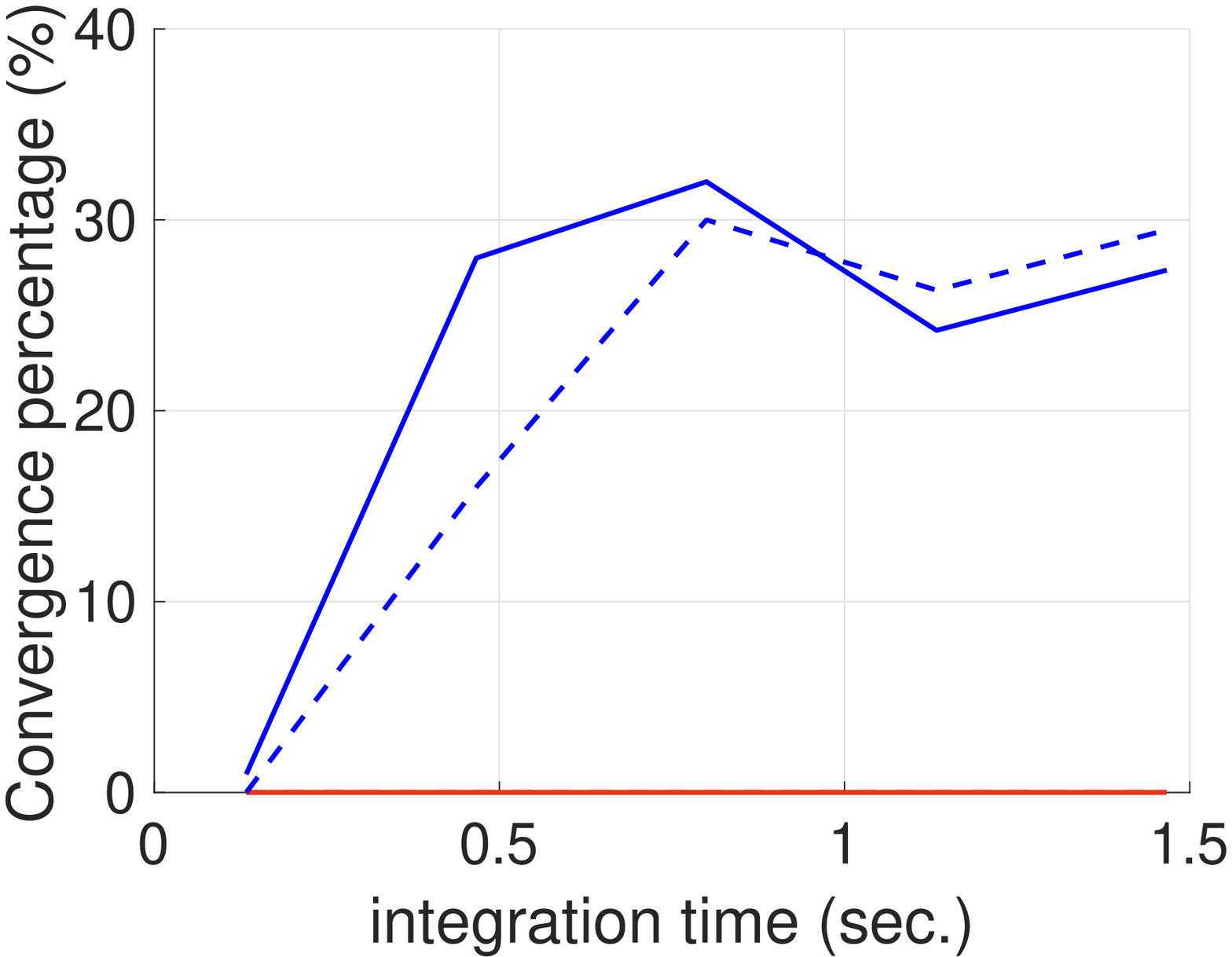} &
         \includegraphics[width=0.33\linewidth]{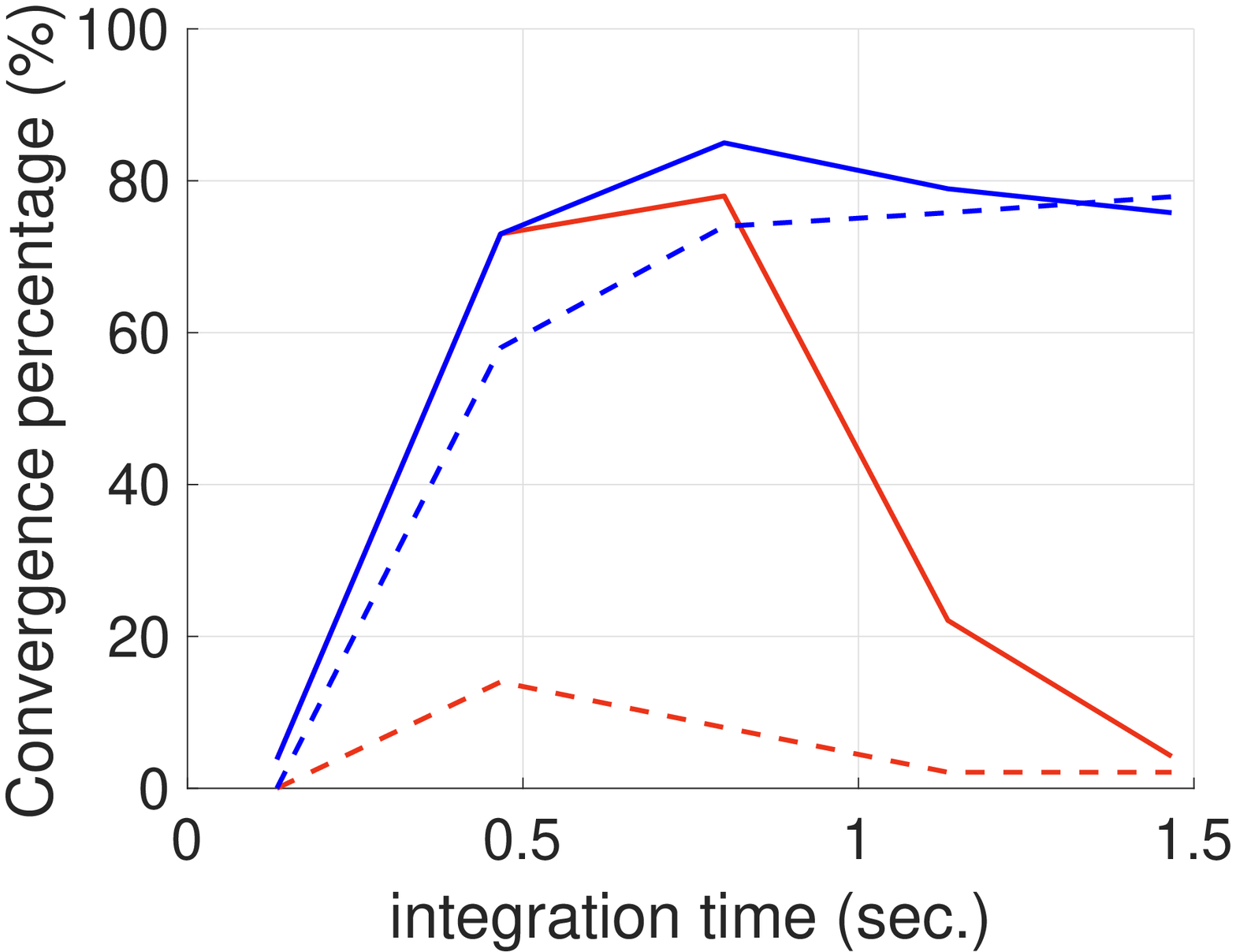} &
        \includegraphics[width=0.33\linewidth]{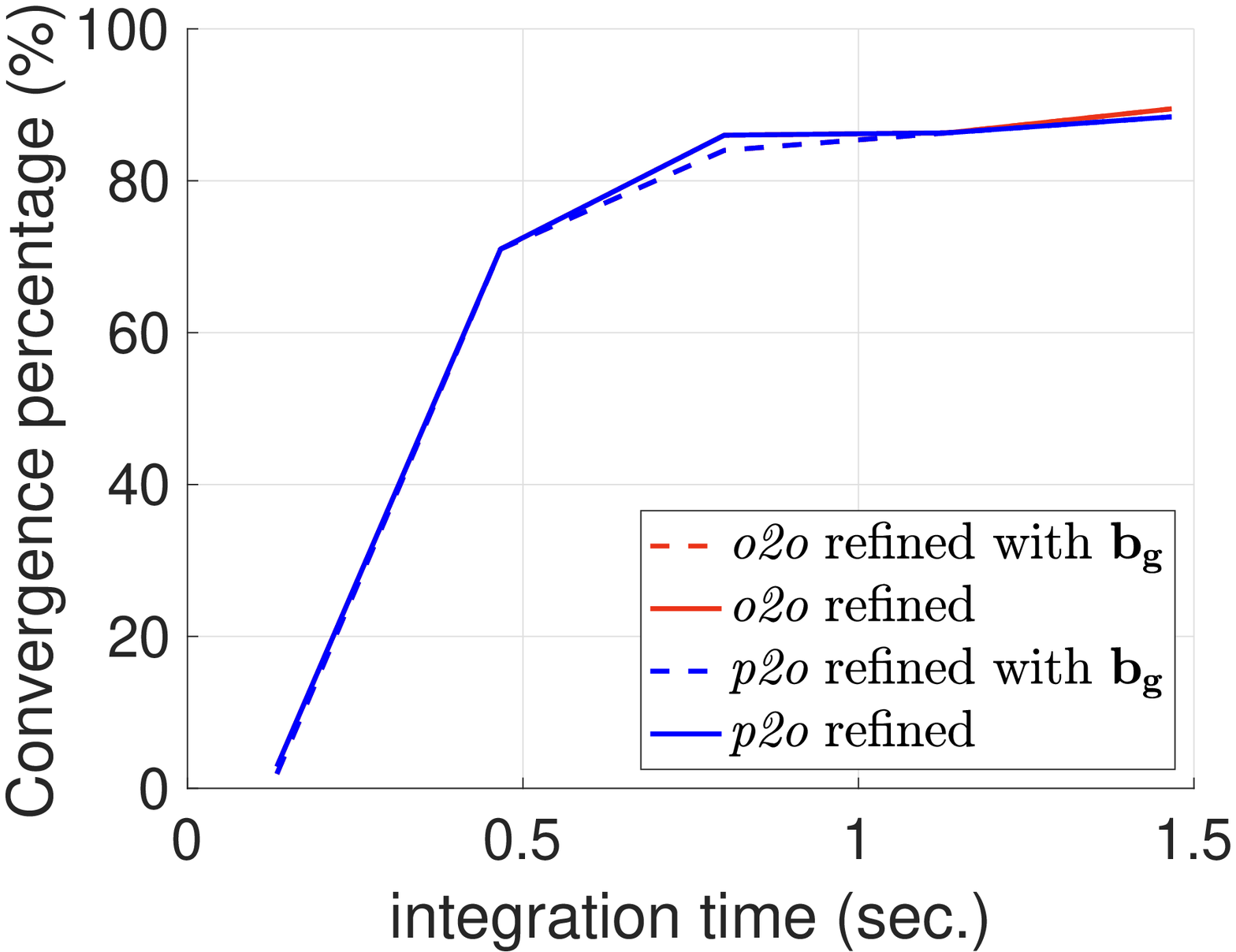} \\
         ~($5$ iterations) & ~($10$ iterations) &~($15$ iterations) \\[1ex]
     \end{tabular} 
    \end{center}
 \caption{Frequency of convergence for non-linear refinement as function of integration time when (a) $5$, (b) $10$ and (c) $15$ iterations are allowed.}
 \label{fig:frequencyOfConvergence}
\end{figure*}

Next, we investigate how the integration time affects the performance. Recall  that the goal is a fast and reliable initialization. We repeat the experiment with $\sigma_u = 0.3$ and test several downsampling factors $N_f$ from $1$ to $9$, which implies the integration time range $[0.13,~1.46]$ seconds. 
Fig.~\ref{fig:performanceWrtIntegrationTime} shows the error as a function of integration time. As expected, the shorter the integration time is, the more sensitive the solvers are. The angle error of gravity estimation, in particular, can reach $3$ degrees at very short integration times. However, it seems that the solvers provide acceptable results after $0.5$ seconds. The proposed solver outperforms and achieves more accurate estimation at any integration time. While the gravity estimation is slightly better, the velocity and the point reconstruction error is decreased by $50\%$  across the whole tested range.  Again, the benefit from enforcing the norm equality constraint on gravity vector is minor. It is noted that one should expect higher integration times when monocular camera is used. We experimentally found here that acceptable estimations are obtained after $1.5$s with a monocular sensor.



As verified by~\cite{Kaiser-RAL2017}, the accelerometer bias, when separable from the gravity, does not affect the closed-form solution. Rather, the gyroscope bias does affect the performance, when its magnitude is relatively large and the integration time is long. We reached similar conclusions for both solvers. Therefore, we model the gyroscope bias along with the non-linear refinement in the next experiment. 

\subsection{Refinement performance evaluation}
We here evaluate the contribution of the closed-form solvers to the non-linear refinement. Out refence is the refiner discussed in Sec.~\ref{sec:nonLinearRefinement} that minimizes the re-projection error. The rotated gravity vector is modelled by \eqref{eq:gravityModel} and the gyroscope bias is optionally modelled. To initialize the structure when \emph{o2o} solver is used, we average the $\lambda$-based reconstructions per point. The analytic Jacobians needed for the optimization are given in Appendix \ref{appendix}.

\begin{table*}[h!]
 \caption{Comparison of average performance per sequence; .}
    \label{tab:realMatchesResults}
  \begin{center}
     \begin{tabular}{|l|c|c|c|} 
         \hline
        &\scriptsize{ \MM{Walking}}  &  \scriptsize{\MM{Running}} & \scriptsize{\MM{HeadMoving}}\\
        & {\scriptsize vel.(\%)/grav.(deg.)} & {\scriptsize vel.(\%)/grav.(deg.)} & {\scriptsize vel.(\%)/grav.(deg.)} \\
      \hline
       \hline
      \emph{o2o} closed-form  &  $2.98\%$ / $0.147^{\circ}$     &   $5.01\%$ / $0.607^{\circ}$       &        $9.82\%$ / $0.216^{\circ}$        \\
      \emph{p2o} closed-form  &  $\bf{2.76}\%$ / $\bf{0.143}^{\circ}$     &   $\bf{2.42}$ / $\bf{0.339}^{\circ}$       &        $\bf{6.52}\%$ / $\bf{0.205}^{\circ}$        \\
        \hline
        \hline
      \emph{o2o} refined   &  $2.77\%$ / $0.141^{\circ}$     &   ${0.45}\%$ / ${0.145}^{\circ}$       &        $5.40\%$ / $0.164^{\circ}$       \\
     \emph{p2o} refined  &  $2.77\%$ / $0.140^{\circ}$     &   ${0.45}\%$ / ${0.145}^{\circ}$       &        $5.40\%$ / $0.163^{\circ}$       \\

      \emph{o2o} refined (Cauchy loss)  &  $2.72\%$ / ${0.124}^{\circ}$      &   $2.96\%$ / $0.313^{\circ}$       &        $4.77\%$ / $0.149^{\circ}$        \\
     \emph{p2o} refined (Cauchy loss)  &  ${2.71}\%$ / $0.125^{\circ}$     &   $0.62\%$ / $0.166^{\circ}$       &        ${4.76}\%$ / ${0.148}^{\circ}$        \\

\hline
 \hline
VIO (Kalman filter)  &  $3.05\%$ / $0.099^{\circ}$     &   $3.84\%$/$0.258^{\circ}$       &        $4.85\%$ / $0.151^{\circ}$        \\     
 \hline
    \end{tabular}
  \end{center}
\end{table*}

A modification of the Levenberg-Marquardt framework of~\cite{Loukaris-ICCV2005} is used for minimization. All the thresholds of the stop criteria in~\cite{Loukaris-ICCV2005} are set to $10^{-9}$ and we let the algorithm terminate. 

Fig.~\ref{fig:nonLinearRefinementPerformance} shows the error of algorithms per iteration, for the case of $N_f=3$ and $\sigma_u = 0.3$, averaged over all the realizations and tested frame windows of the sequence. 
Notably, when starting from \emph{o2o} solution, more than five iterations are needed to just reach the accuracy of the \emph{p2o} solver. As a result, similar accuracy can be achieved with much less operations. All the counterparts reach almost comparable floor values after $12$ iterations which implies a locally convex error function. When the gyroscope bias is modelled and estimated, further non-linearities are introduced and the convergence may be slower. It is noted that the rate of convergence remained unaffected after adjusting the initial damping factor. 

We also compare the minimizers in terms of the frequency of convergence. In a real scenario, one would allow a few iterations while he would be more interested in IMU state initialization (the points may be re-triangulated after initialization). Therefore, we consider that the algorithm has converged after a predefined number of iterations when the relative velocity error is below $0.025$ \emph{and} the angular error is below $0.25$ degrees. Then, we use this criterion to count the successful realizations for a specific number of iterations.
 Fig.~\ref{fig:frequencyOfConvergence} shows the percentage of convergence as a function of integration time for $5$, $10$ and $15$ iterations. Unlike \emph{p2o} solver, the \emph{o2o} solver would most likely fail to well initialize the state unless a sufficient number of iteration is allowed. We noticed that the vast majority of Levenberg-Marquardt iterations includes a single cost-function evaluation.



Although the current IMU readings include a time varying accelerometer and gyroscope bias, we did not observe any improvement due to their modelling. This is most likely because of the low noise levels of the used device compared to the dominant tracking error. We experimentally confirmed the ability of the algorithms to estimate high yet unrealistic biases that were artificially added. 

When an iterative optimizer is used, another approach is to use the unconstrained linear solver to get the initial velocity and gravity, and then let the optimizer refine the parameters and estimate the biases.




\subsection{Real correspondences}
Different trajectories and types of motion are here combined with different $3D$ scenes. In all the sequences, the rig is static at the beginning. A rotation-aware ECC-based tracker on FAST corners tracks a maximum number of $200$ points per rendered image. As a reference baseline, a well initialized (due to the static part) extended Kalman filter that propagates IMU states
using visual and inertial data is also employed~\cite{Mourikis-ICRA2007}. Again, $5$ downsampled frames with $N_f=3$ are used per frame window. To compensate for mismatches and non-Gaussian tracking error, a Cauchy loss~\cite{Hartley-Book2004} in \eqref{eq:totalError} is also tested. A maximum number of $15$ iterations for the refinement is allowed.

\begin{figure*}[t]
  \begin{center}
      \begin{tabular}{@{\hspace{0mm}}c@{\hspace{2mm}}c@{\hspace{2mm}}c}
         \includegraphics[width=0.33\linewidth]{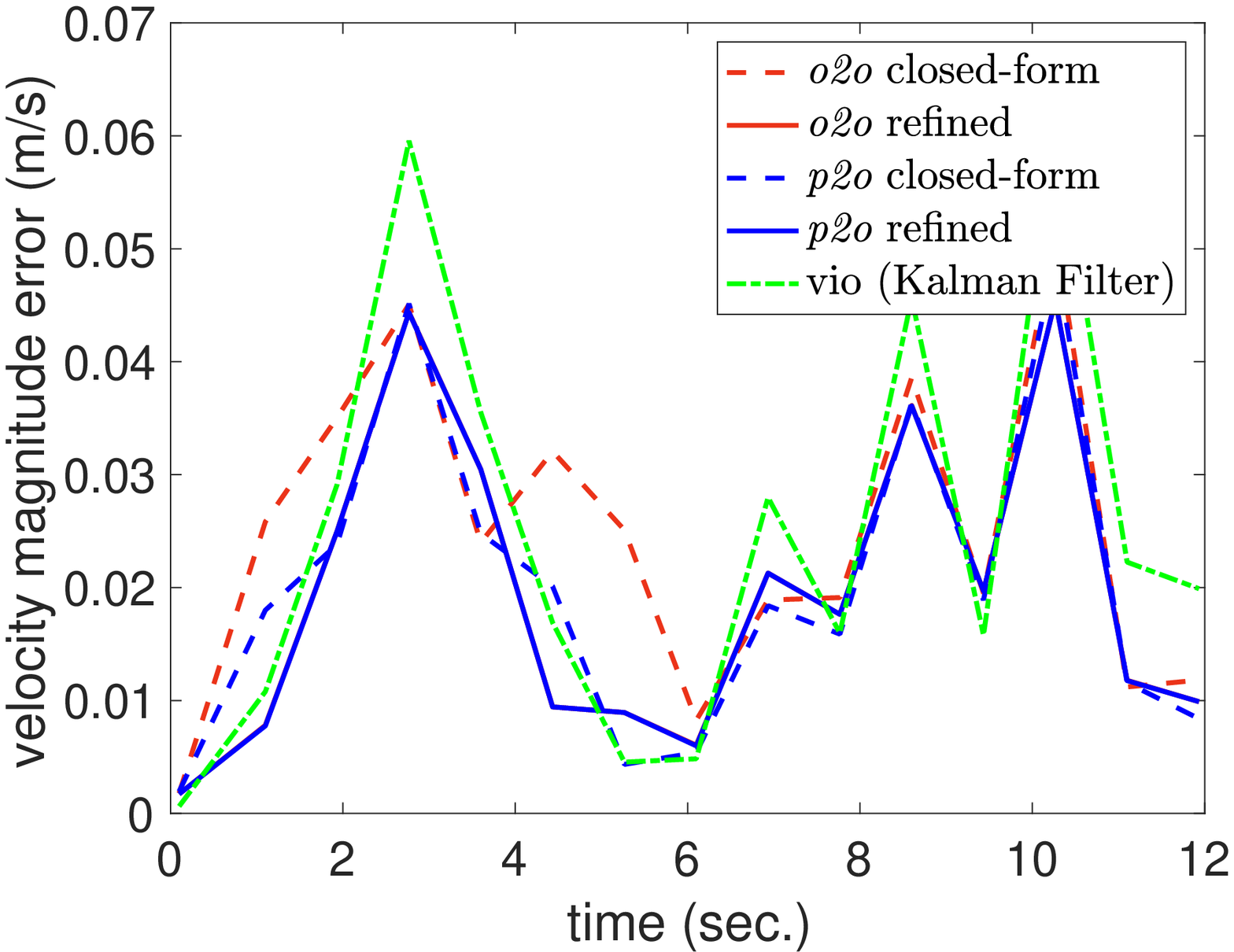} &
                  \includegraphics[width=0.33\linewidth]{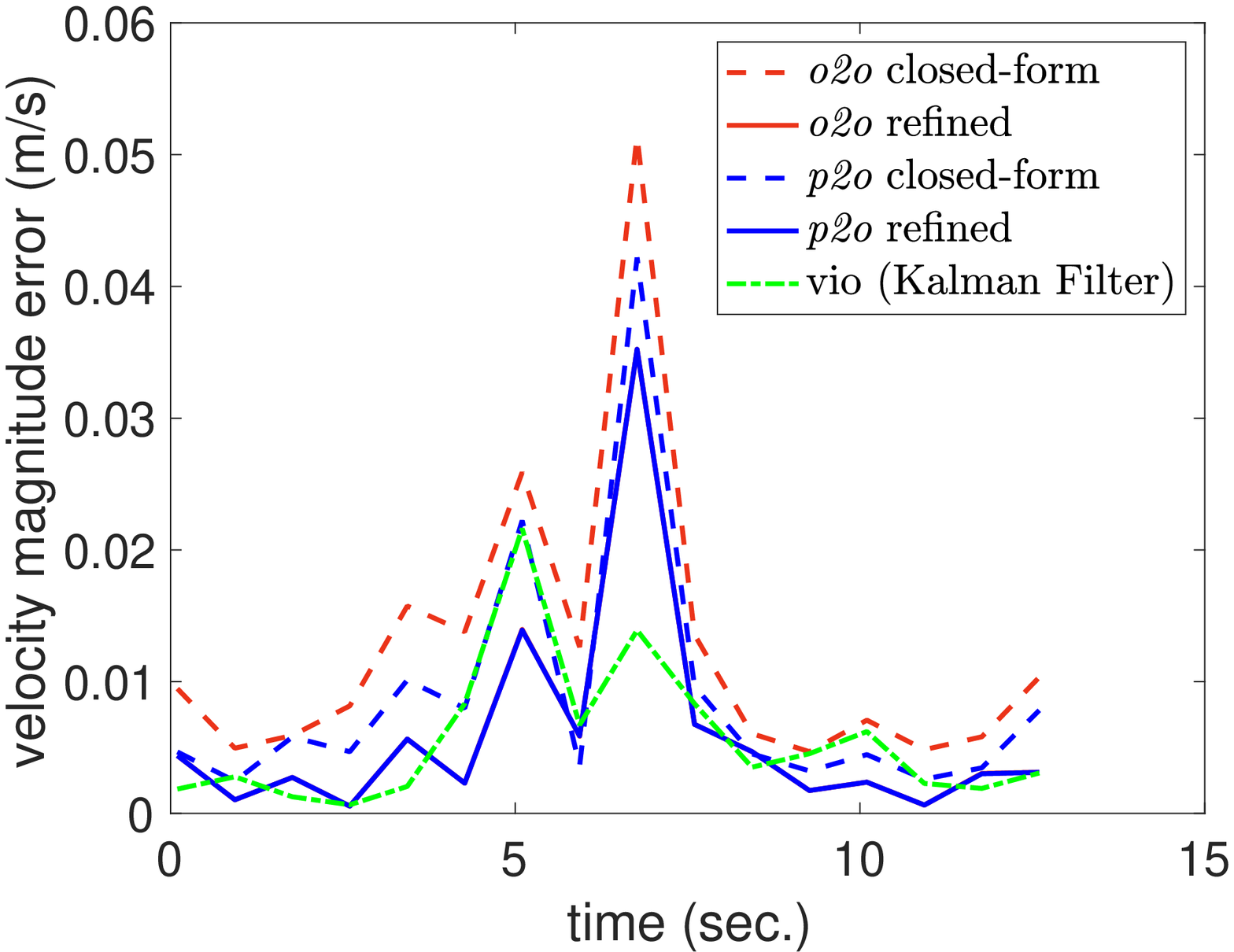} &
                           \includegraphics[width=0.33\linewidth]{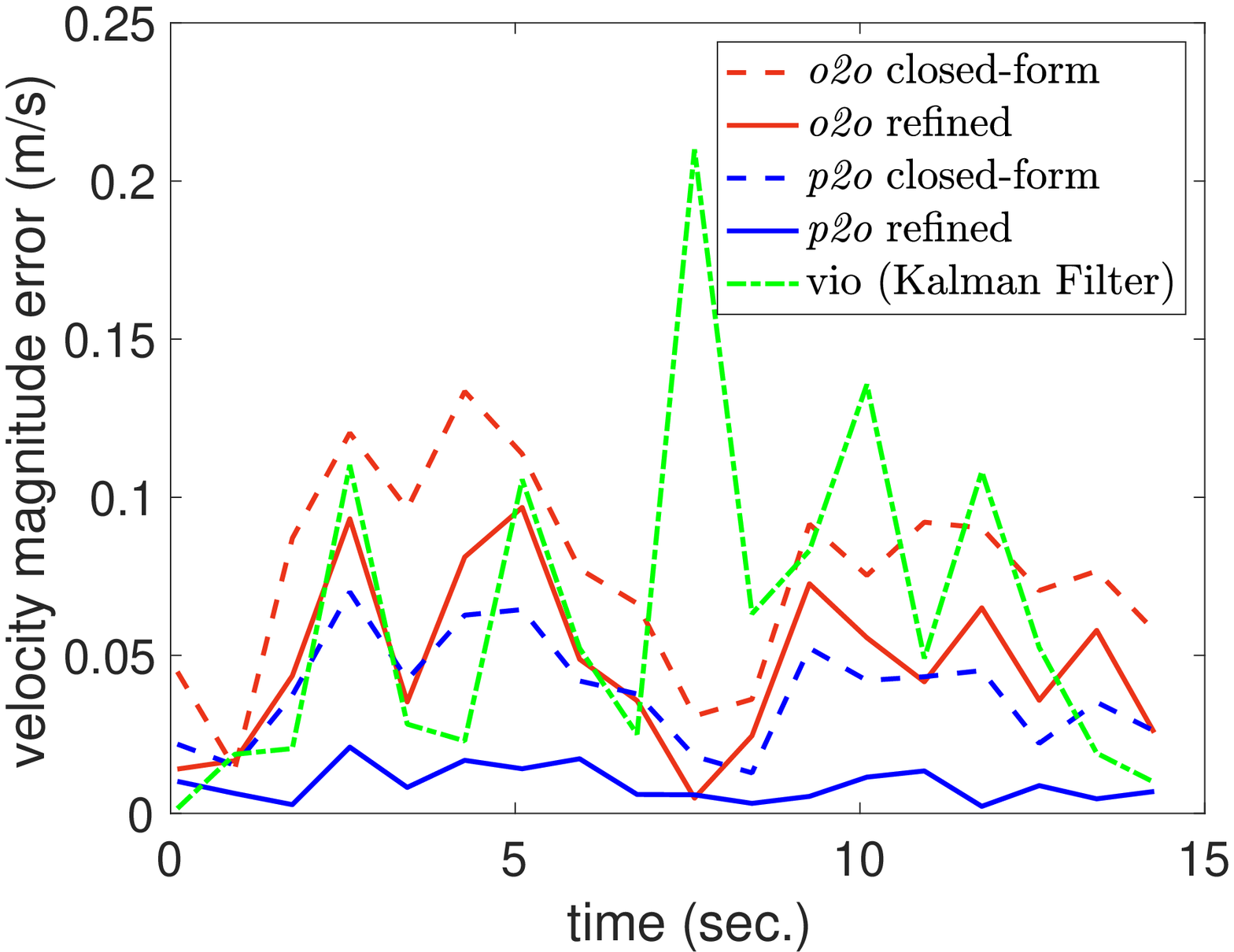}\\ 
\\
        \includegraphics[width=0.33\linewidth]{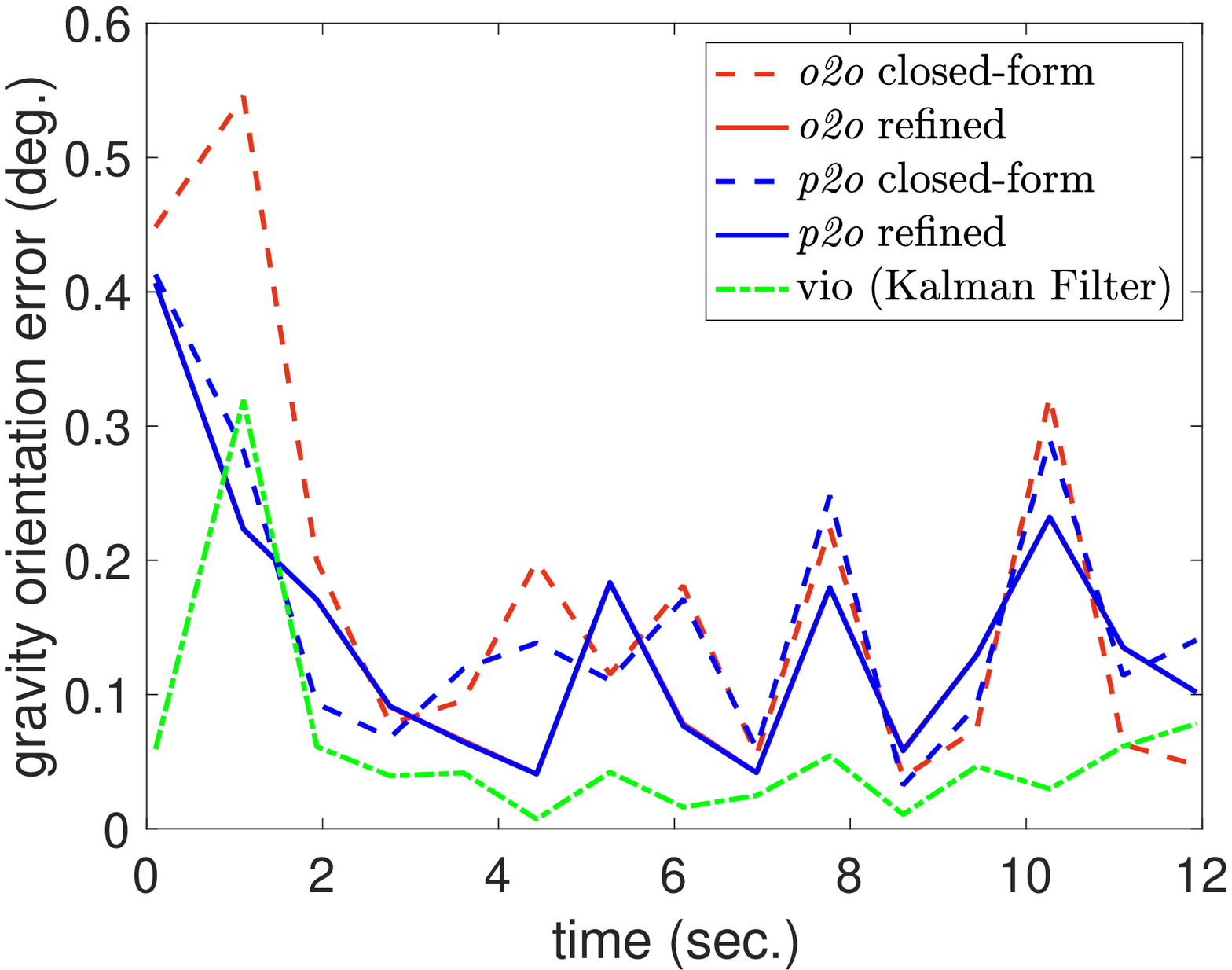} &
        \includegraphics[width=0.33\linewidth]{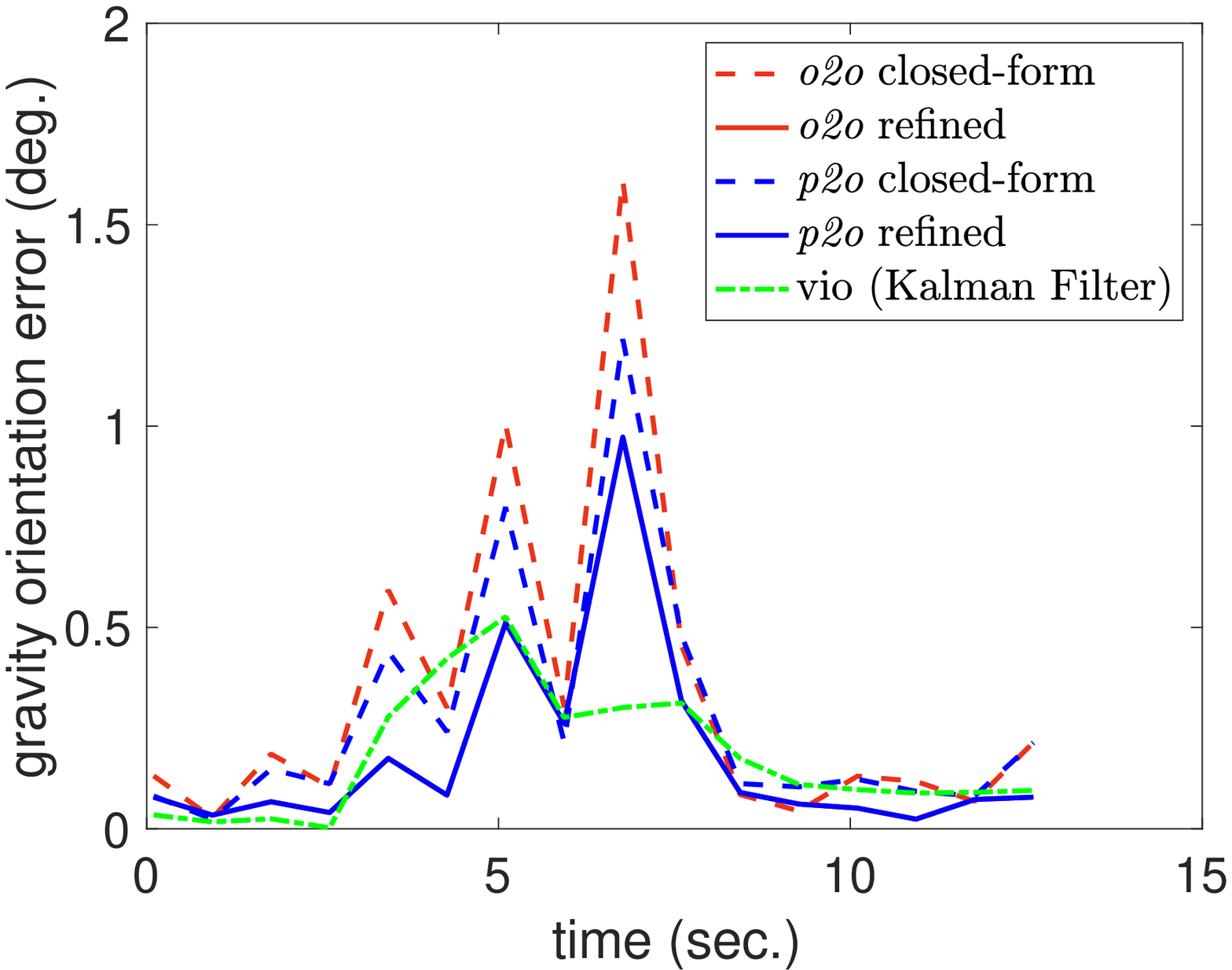} &
        \includegraphics[width=0.33\linewidth]{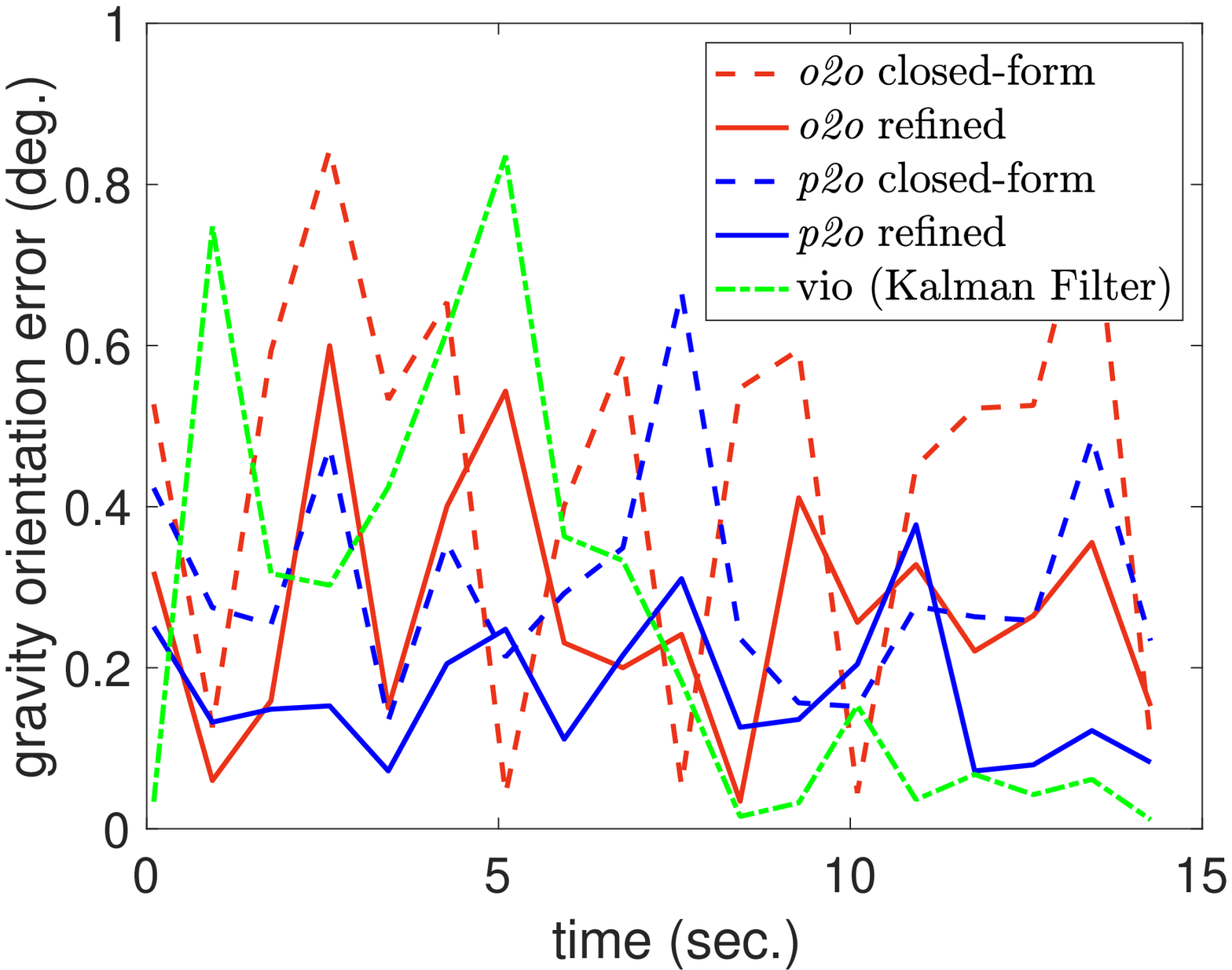} \\
\footnotesize{\MM{Walking}, maximum velocity: $0.96$m/s}&
\footnotesize{\MM{HeadMoving}, maximum velocity: $0.94$m/s} &
\footnotesize{ \MM{Running}, maximum velocity: $2.80$m/s} \\
        
     \end{tabular} 
    \end{center}
 \caption{(\emph{top}) Velocity and (\emph{bottom}) gravity orientation error per frame; point correspondences are delivered by a feature tracker on rendered images; the integration time is $0.46$s.}
 \label{fig:realMatchesPerformance}
\end{figure*}

We test the algorithms on three sequences, \MM{Walking}, \MM{HeadMoving} and \MM{Running}. Table~\ref{tab:realMatchesResults} summarizes the average error over all frame windows per sequence.  Overall, the \emph{p2o} solver obtains better estimates than the \emph{o2o} solver. When a non-linear refiner follows, the error further decreases. Note that Cauchy loss may lead to slower convergence and a larger error may be achieved for a small number of iterations. Fig.~\ref{fig:realMatchesPerformance} shows the error over time. The absolute velocity magnitude error is here shown instead, and the maximum velocity is also given. Interestingly, the velocity estimation of the closed-form is comparable with the one from Kalman filter for the \MM{Walking} and \MM{HeadMoving} sequences. 
The \MM{Running} sequence is more challenging because of jumping while jogging. The proposed solver combined with the refiner clearly outperforms in this case. 


\subsection{Real data}
The above experimental setup regards rendered image sequences from artificial 3D scenes. We here focus on the the closed-form solvers and test them on two real sequences acquired from Snap Spectacles in a typical open office space. The first sequence, named \MM{OfficeLoop}, is a 25m loop-shaped walking sequence within the office. The second sequence, named  \MM{LookingAround}, is more challenging and has rotational motion with strong velocity variation where the wearer looks around without stepping. In both cases, the wearer is static in the beginning as well as in the end of the recording. 

\begin{figure*}[t]
  \begin{center}
      \begin{tabular}{@{\hspace{0mm}}c@{\hspace{2mm}}c@{\hspace{2mm}}c}
         \includegraphics[width=0.33\linewidth]{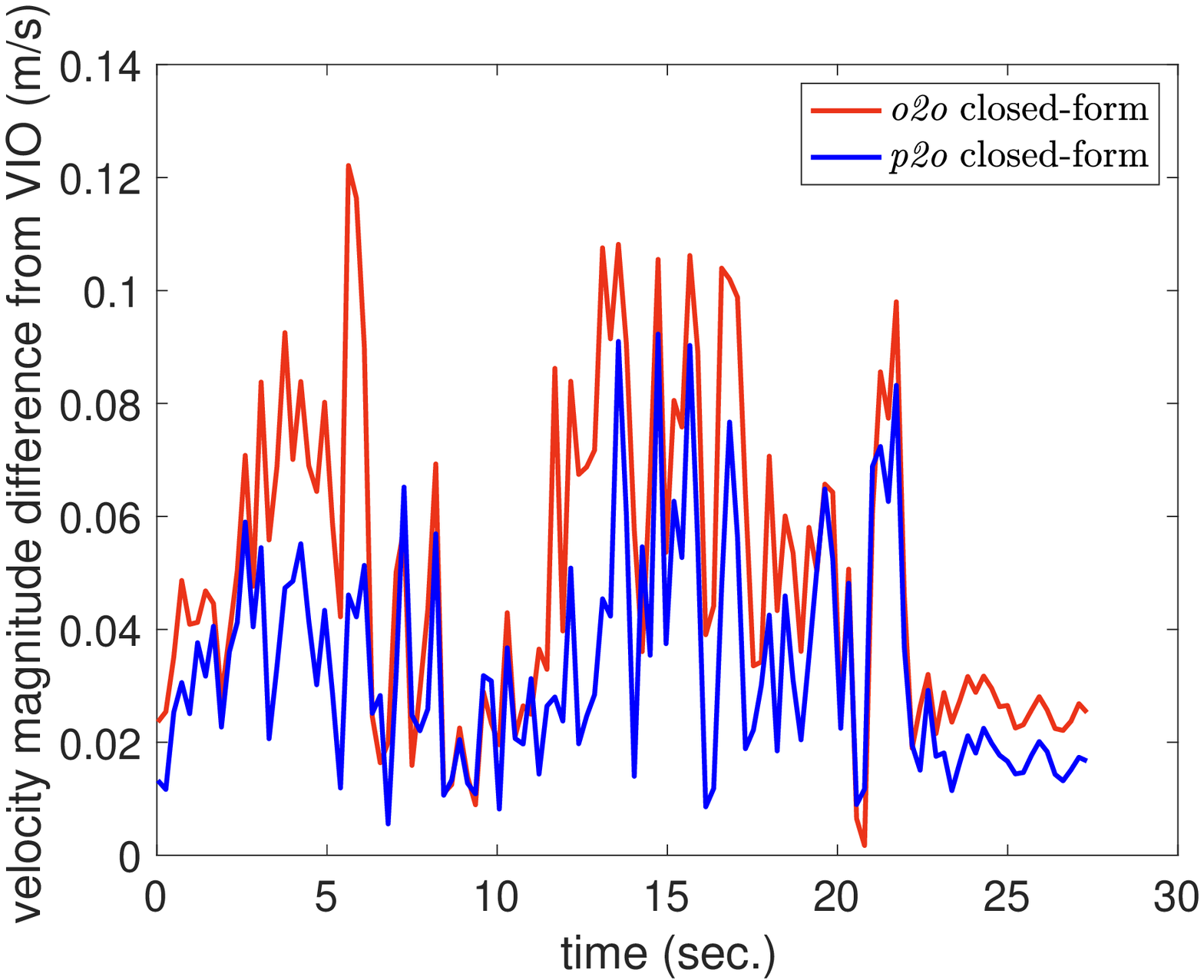} &
        \includegraphics[width=0.33\linewidth]{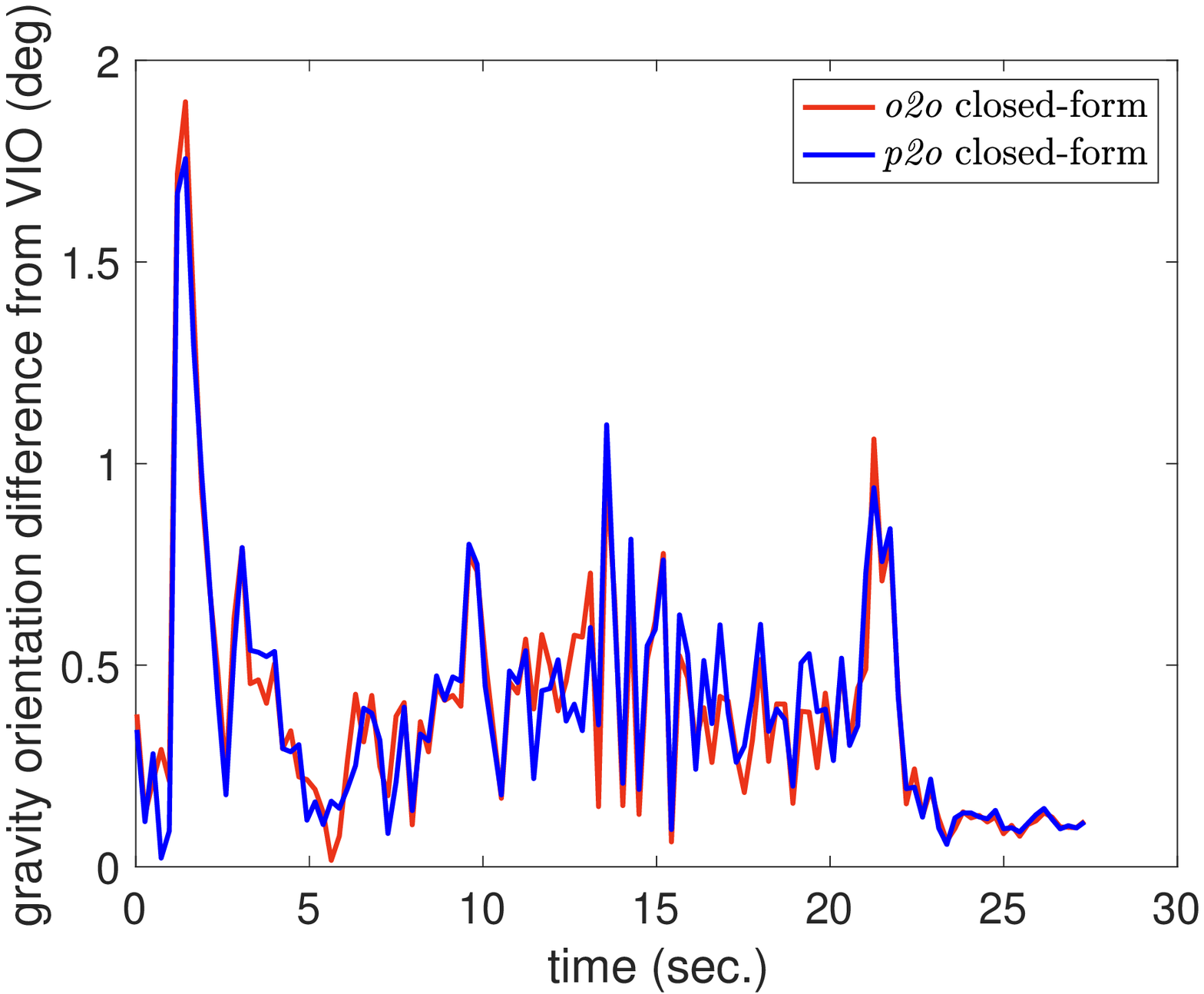} & 
        \includegraphics[width=0.33\linewidth]{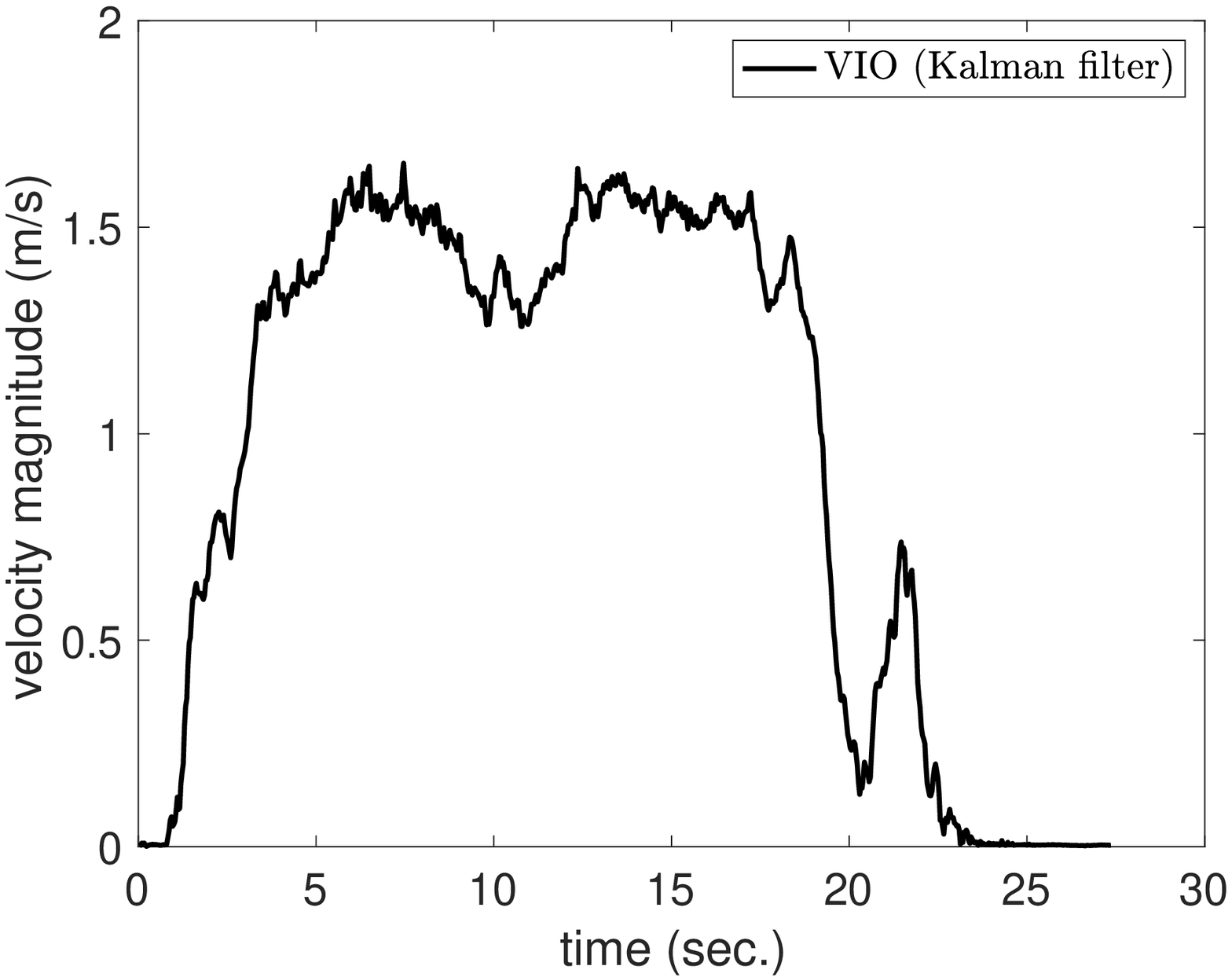}   \\
        \multicolumn{3}{c}{\MM{OfficeLoop}} \\[1ex]
         \includegraphics[width=0.33\linewidth]{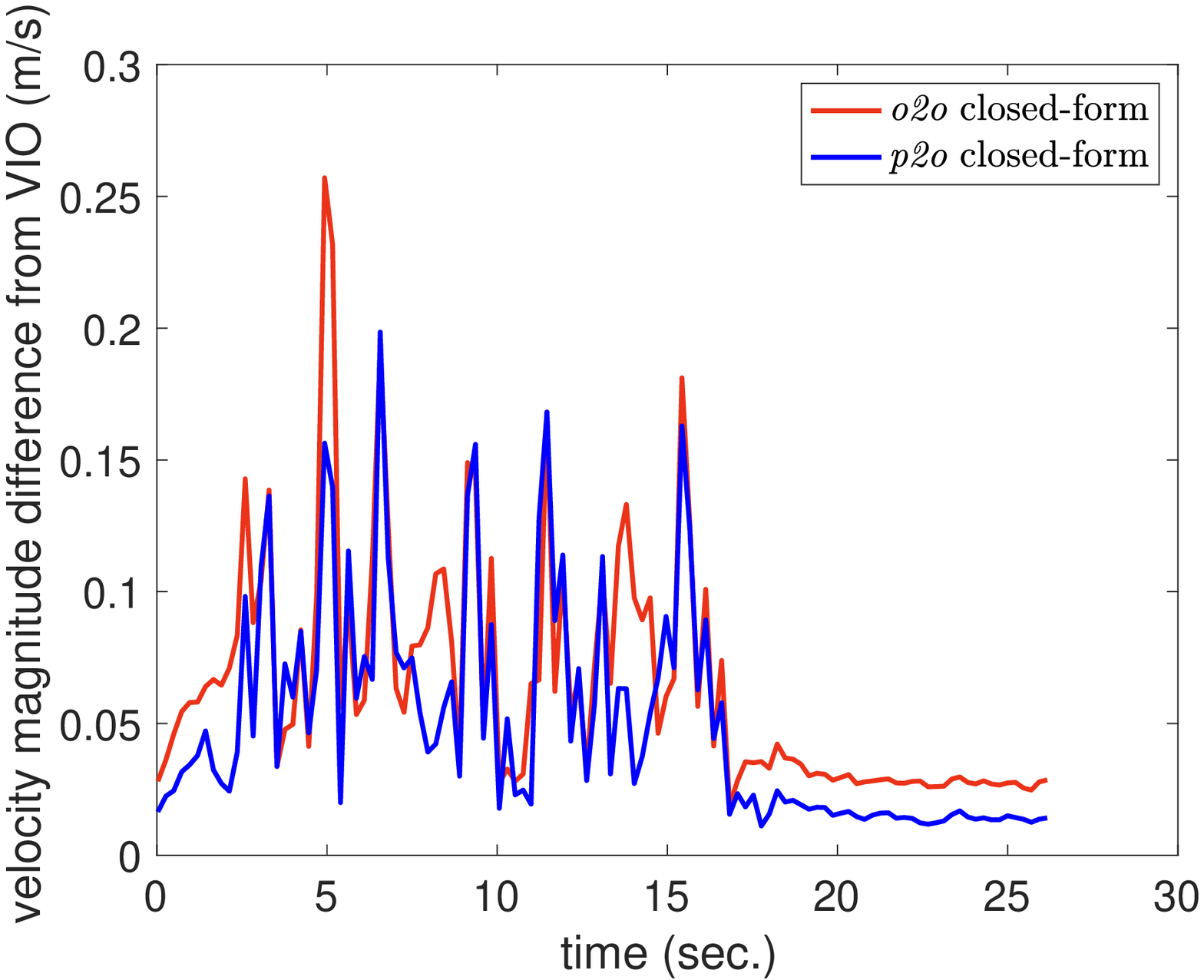} &
        \includegraphics[width=0.33\linewidth]{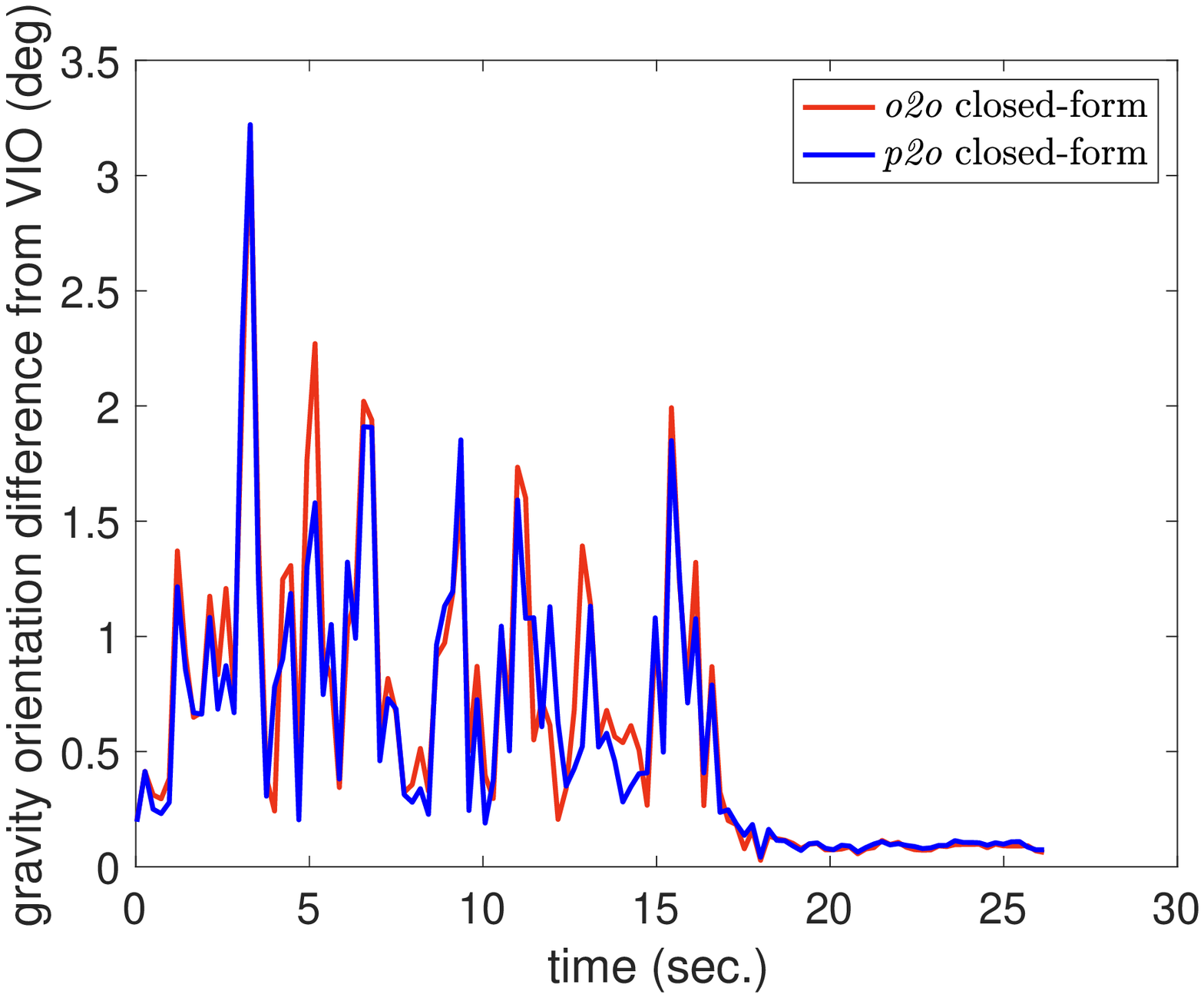} &
        \includegraphics[width=0.33\linewidth]{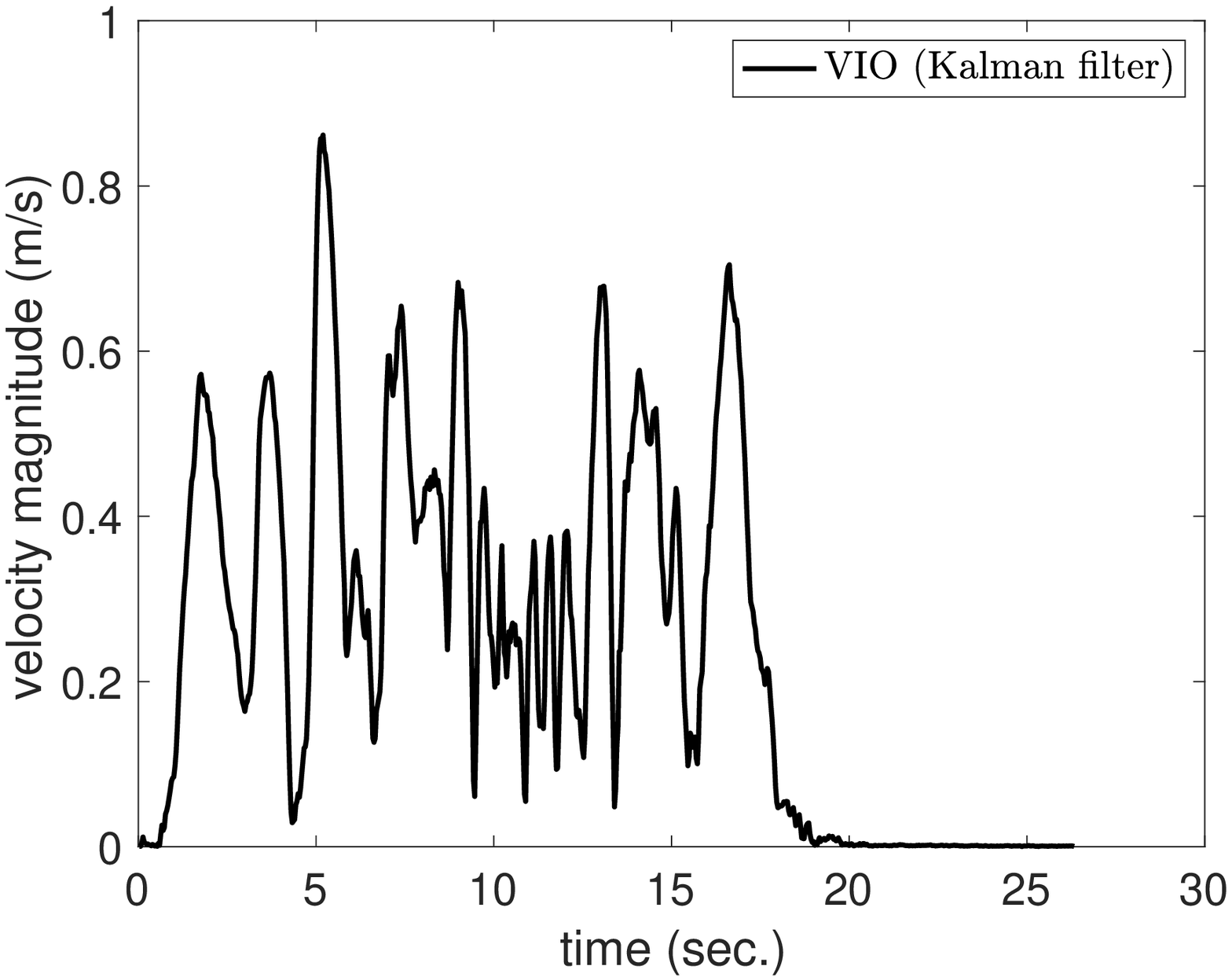}   \\
         \multicolumn{3}{c}{\MM{LookingAround}} \\[1ex]  
     \end{tabular} 
    \end{center}
\caption{(\emph{left}) Velocity and (\emph{middle}) gravity orientation difference from VIO (VIO velocity is shown on the \emph{right} side ) after testing the closed-form solvers on real data delivered by Snap Spectacles.}
 \label{fig:realDataPerformance}
\end{figure*}

We employ the closed-form solvers and re-initialize the state of every frame using a moving $7$-frame window of downsampled frames ($N_f=3$). Since the ground truth is not available, we show the deviation from the VIO  baseline which uses prior information for the state estimation. The velocity and orientation differences are shown in Fig.~\ref{fig:realDataPerformance}. Although the two solvers provide similar gravity orientations, the velocity estimations are quite different. Unlike the proposed {\it p2o} solver, the velocity estimations of the {\it o2o} solver are quite far from VIO velocities. As expected, the estimations for the \MM{LookingAround} sequence are worse due to the rapid velocity and rotation changes, which in turn make the feature tracking break more often. 

\subsubsection*{\bf{Time comparison}}
Both closed-form solvers were implemented with Eigen C++ library. A sparse linear system and solver are used for \emph{o2o}. Instead, the block structure of matrices in (\ref{eq:totalLinearSystemEliminatedTwice}) enables the direct construction of the least-squares solution system. That is, for the unbiased case, we start with a $6\times 6$ zero matrix and $6\times 1$ zero vector and we update them after processing each track (point). 
Such an implementation is memory efficient too, that is, the use of large (sparse) matrices is not necessary. 
Given IMU data and the visual tracks, the average build-and-solve times over all tested frame windows are shown in Table \ref{tab:executionTimes}. A $4\times$ faster initialization is achieved for a $7$-frame window. As a consequence, there is a substantial gain from replacing \emph{o2o} with \emph{p2o} even when the performance is comparable.

\begin{table}[b!]
 \caption{Comparison of average times per sequence.}\vskip -0.2cm
    \label{tab:executionTimes}
 \begin{center}
  \scriptsize
     \begin{tabular}{|l|c|c|} 
         \hline
         & \multicolumn{2}{c|}{average running times (C++, i7/2.6GHz)}\\
\cline{2-3}
        &\scriptsize{ \MM{OfficeLoop}}  &  \scriptsize{\MM{LookingAround}}\\
      \hline
       \hline  
      \scriptsize\emph{o2o} closed-form  &  $4.27$ms      &   $7.50$ms     \\
             \hline  
      \scriptsize\emph{p2o} closed-form  &  $\bf{1.18}$ms    &   $\bf{1.96}$ms \\    
 \hline
    \end{tabular}\\
  \end{center}\vskip -0.2cm
\end{table}


\section{Conclusions}\label{sec:Conclusions}
A new closed-form solver for the vi-SfM problem was suggested, in the context of VIO and SLAM initialization. The mathematical derivation along with the experimental validation show that the solver is more accurate and faster than the state-of-the-art. Either as a stand-alone solver or combined with a non-linear optimizer that further refines the initial state, it offers a significant speedup to the initialization phase of VIO and SLAM pipelines.

%
%
\bibliographystyle{IEEEtranS}
\bibliography{main}

\begin{appendices}

\appendix
\subsection{Jacobians}
\label{appendix}

In this section, we provide the analytic Jacobians for the miminization of $f(\V{x})$  in \eqref{eq:totalError}. For the sake of simplicity, we assume a squared Euclidean distance $d$, and a single point 
$\V{m}$ to skip the index $j$. Recall the importance of index $i$ when multiple points are used with a rolling-shutter camera,  that is, each observation is captured at different time, and from a different viewpoint. 

Suppose a state parameter vector $\V{x}$ obtained from the closed-form solution. The goal of the refiner is to find a correction $\Delta \V{x}$ such that $f(\V{x}+\Delta \V{x}) < f(\V{x})$. If we express the reconstructed point at the camera of the $i$-th timestamp as $\V{w}_{i} = \M{R}_{C_{i}}^{^{\top}}\V{m} - \M{R}_{C_{i}}^{^{\top}}\V{p}_{C_{i}}$, the linearized problem is written as
\begin{equation}
\min_{\Delta\V{x}} \sum_{i} \|\pi(\V{w}_{i}) - \pi(\V{u}_{i}) + \M{J}_{i}\Delta\V{x}\|^2
\end{equation}
where $\M{J}_{i}$  is the Jacobian of $\pi(\V{w}_{i})$ \wrt the state parameters $\V{x}$. Note that in the general case of multiple points ($\V{m}_j$, $j=1,\dots,M$), $\M{J}_i$ is replaced by $\M{J}_{ji}$, which is a $2\times (12+3M)$ sparse matrix with  $(4+M)$ blocks of size $2\times 3$. When gravity is modelled by \eqref{eq:gravityModel}, $\M{J}_{ji}$ is of size $2\times (11+3M)$ and the gravity Jacobian block has size $2\times 2$.  Only the first four blocks as well the $(4+j)$-th block have non-zero elements per observation.

We below provide the five blocks of the Jacobian $\M{J}_{i}$.  All the blocks include the Jacobian $\M{J}_{\pi}$ of the projection operator $\pi(\V{w}_{i})$. If we assume a $3D$ vector $\V{w}_{i}=[x_{i}, y_{i}, z_{i}]^{\top}$ and the projection operator $\pi(\V{w}_{i})= [x_{i}/z_{i}, y_{i}/z_{i}]^{\top}$, its Jacobian is given by
\begin{equation}\label{eq:projectionJacobian}
\M{J}_{\pi} = \frac{1}{z_{i}}
\left[ 
\begin{array}{cc}
    \M{I}_2 & ~ -\pi(\V{w}_{i}) \\
    \end{array}
  \right].
\end{equation}
The first block of  $\M{J}_{i}$ regards the velocity and is given by:
\begin{equation}\label{eq:velocityJacobian}
\frac{\partial \pi(\V{w}_{i})}{\partial \V{v}_0} = -t_{i}\M{J}_{\pi}\M{R}_{C_{i}}^{^{\top}}\
\end{equation}
The second block of  $\M{J}_{i}$ regards the gravity and in case that the norm constraint is not enforced is simply given by :
\begin{equation}\label{eq:gravityJacobian}
\frac{\partial \pi(\V{w}_{i}) }{\partial \V{g}_0} = -\frac{t_{i}^2}{2}\M{J}_{\pi}\M{R}_{C_{i}}^{^{\top}}.
\end{equation}
When the gravity is modelled by \eqref{eq:gravityModel}, this block becomes $2\times 2$ and the two columns are given by
\begin{equation}\label{eq:angleXjacobian}
\frac{\partial \pi(\V{w}_{i}) }{\partial \phi_x} = \frac{\partial \pi(\V{w}_{i}) }{\partial \V{g}_0}\left(\gamma_c \left[\begin{array}{c} \phi_x\phi_y \\ -\phi_x^2 \\ 0 \end{array}\right] + \gamma_s \left[\begin{array}{c} -\phi_x\phi_y \\ \phi_x^2 - \|\boldsymbol{\phi}\|^2 \\ -\phi_x \|\boldsymbol{\phi}\|^2 \end{array}\right]\right)
\end{equation}
and
\begin{equation}\label{eq:angleYjacobian}
\frac{\partial \pi(\V{w}_{i}) }{\partial \phi_y} =\frac{\partial \pi(\V{w}_{i}) }{\partial \V{g}_0}\left( \gamma_c \left[\begin{array}{c} \phi_y^2 \\  -\phi_x\phi_y \\ 0 \end{array}\right] + \gamma_s \left[\begin{array}{c} \|\boldsymbol{\phi}\|^2 - \phi_y^2 \\ \phi_x\phi_y \\  -\phi_y \|\boldsymbol{\phi}\|^2 \end{array}\right]\right),
\end{equation}
where $\gamma_c = \frac{\gamma\cos(\| \boldsymbol{\phi} \|)}{\| \boldsymbol{\phi} \|^2}$ and $\gamma_s = \frac{\gamma\sin(\| \boldsymbol{\phi} \|)}{\| \boldsymbol{\phi} \|^3}$.
The third block of  $\M{J}_{i}$ regards the accelerometer bias and is given by:
\begin{equation}\label{eq:accBiasJacobian}
\frac{\partial \pi(\V{w}_{i}) }{\partial \V{b}_a} = -\M{J}_{\pi}\M{B}_{i}\M{R}_{C_{i}}^{^{\top}}\
\end{equation}
The fourth block of  $\M{J}_{i}$ regards the gyroscope bias and is approximated by:
\begin{dmath}\label{eq:gyroBiasJacobian}
\frac{\partial \pi(\V{w}_{i}) }{\partial \V{b}_g} \simeq -\M{J}_{\pi} {\M{R}_{C}^{I}}^\top \left( [\M{R}_{I_{i}}^{\top}\V{m}]_\times \frac{\partial \M{R}_{I_{i}}}{\partial\V{b}_g } +  \sum_{k=0}^{j-1} \beta_{ki} [\M{R}_{I_{i}}^{\top}\M{R}_{I_{k}}\V{a}_k]_\times \left(\frac{\partial \M{R}_{I_{i}}}{\partial\V{b}_g } - \frac{\partial \M{R}_{I_{k}}}{\partial\V{b}_g } \right)\right)
\end{dmath}
where $[.]_\times$ denotes the skew-symmetric matrix and $\frac{\partial \M{R}_{I_{i}}}{\partial\V{b}_g }$ is the Jacobian of the rotation \wrt the gyroscope bias, approximated  by,
\begin{equation}\label{eq:rotationJacobian}
\frac{\partial \M{R}_{I_{i}}}{\partial\V{b}_g }  \simeq \M{R}_{I_{i}}^{\top} \sum_{k=0}^{n_i-1}{\M{R}_{I_{k+1}} \M{\Omega}_k T_s},
\end{equation}
where $\M{\Omega}_k$ is the \emph{right Jacobian} of SO3 at $\boldsymbol{\omega}_k$~(\cite{Forster-TRO2017}, Eq.8). The computation of $\frac{\partial \V{w}_{i}}{\partial \V{b}_g}$ and $\frac{\partial \M{R}_{I_{i}}}{\partial\V{b}_g }$ stems from the properties of exponential map~\cite{Forster-TRO2017}. The fifth Jacobian that regards the point is simply given by
\begin{equation}\label{eq:pointJacobian}
\frac{\partial \pi(\V{w}_{i}) }{\partial \V{m}} = \M{J}_{\pi}\M{R}_{C_{i}}^{^{\top}}\
\end{equation}

\end{appendices}

\end{document}